%% file: acl_latex.tex
\pdfoutput=1

\PassOptionsToPackage{svgnames,dvipsnames}{xcolor}
\documentclass[11pt]{article}

\usepackage[final]{acl}

\usepackage{times}
\usepackage{latexsym}

\usepackage[T1]{fontenc}

\usepackage[utf8]{inputenc}

\usepackage{microtype}

\usepackage{inconsolata}

\usepackage{graphicx}

\usepackage{
    amsfonts, amsmath, paralist, url, wrapfig, amssymb, titlesec,
    bm, enumitem, longtable, multirow, lipsum, booktabs, subcaption, stfloats, tabularx, soul, makecell, listings, colortbl, paralist, array
}
\usepackage[capitalise]{cleveref}
\usepackage[most]{tcolorbox}

\newcommand{\abr}[1]{\textsc{#1}}
\newcommand{\lda}[0]{\abr{LDA}}
\newcommand{\mallet}[0]{\abr{Mallet}}
\newcommand{\ctm}[0]{\abr{CTM}}
\newcommand{\bertopic}[0]{\abr{BERTopic}}
\newcommand{\llamaThreeOneEightB}[0]{Llama-3.1-8B}
\newcommand{\qwenThreeEightB}[0]{Qwen-3-8B}
\newcommand{\qwenThreeThirtyTwoB}[0]{Qwen-3-32B}
\newcommand{\llamaThreeThreeSeventyB}[0]{Llama-3.3-70B}
\newcommand{\qwenTwoFiveSeventyTwoB}[0]{Qwen-2.5-72B}
\newcommand{\gptFourO}[0]{GPT-4o}

\newcommand{\fittauHTM}[0]{\textsc{fit–}\tau_{\ h:tm}}
\newcommand{\fittauLMTM}[0]{\textsc{fit–}\tau_{\ lm:tm}}
\newcommand{\ranktauHTM}[0]{\textsc{rank–}\tau_{\ h:tm}}
\newcommand{\ranktauLMTM}[0]{\textsc{rank–}\tau_{\ lm:tm}}

\newcommand{\proxann}[0]{\abr{ProxAnn}}

\definecolor{customblue}{RGB}{0, 84, 168}
\definecolor{promptBlue}{RGB}{186,225,255} 
\definecolor{promptGreen}{RGB}{186,255,201}   
\definecolor{promptRed}{RGB}{255,179,186}     

\definecolor{lgray}{RGB}{176, 179, 184}

\newtcolorbox[list inside=prompt,auto counter,number within=section]{prompt}[1][]{
    colbacktitle=black!60,
    fonttitle=\small,
    coltitle=white,
    fontupper=\footnotesize,
    boxsep=3pt,
    left=0pt,
    right=0pt,
    top=0pt,
    bottom=0pt,
    boxrule=1pt,
    #1,
}

\title{\textsc{ProxAnn}:\\Use-Oriented Evaluations of Topic Models and Document Clustering}

\author{Alexander Hoyle\thanks{Equal contribution.} \\
  ETH Z\"urich \\
  \texttt{alexander.hoyle@ai.ethz.ch} \\\And
  Lorena Calvo-Bartolom\'e\footnotemark[1] \\
  Universidad Carlos III de Madrid \\
  \texttt{lcalvo@pa.uc3m.es} \\ \\\AND
  Jordan Boyd-Graber \\
  University of Maryland \\
  \texttt{jbg@umiacs.umd.edu} 
  \\\And
  Philip Resnik \\
  University of Maryland \\
  \texttt{resnik@umd.edu} 
  \\}

\begin{document}
\maketitle

\begin{abstract}

Topic model and document-clustering evaluations either use automated metrics that align poorly with human preferences or require expert labels that are intractable to scale.
We design a scalable human evaluation protocol and a corresponding automated approximation that reflect practitioners' real-world usage of models.
Annotators---or an LLM-based proxy---review text items assigned to a topic or cluster, infer a category for the group, then apply that category to other documents.
Using this protocol, we collect extensive crowdworker annotations of outputs from a diverse set of topic models on two datasets.
We then use these annotations to validate automated proxies, finding that the best LLM proxies are statistically indistinguishable from a human annotator and can therefore serve as a reasonable substitute in automated evaluations.\footnote{\url{https://github.com/ahoho/proxann} contains all human and LLM annotation data, as well as a package (and web interface) to compute metrics on new outputs.}

\end{abstract}

\begin{figure}[ht]
  \centering
  \includegraphics[width=0.48\textwidth]{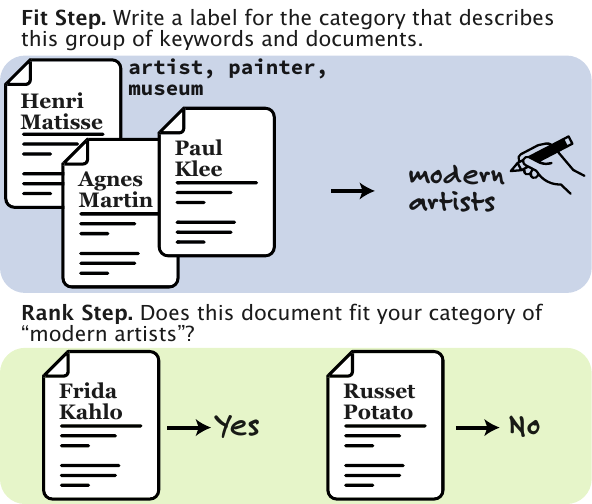}
  \caption{Our evaluation protocol for topic models and document clustering methods. First, a user reviews documents and keywords related to a topic or cluster and identifies a category. Then, they apply that category to new documents (a third ranking step is not shown). The more human relevance judgments align with corresponding model estimates, the better the model. Importantly, the protocol is straightforward to adapt to an LLM prompt, creating a ``proxy annotator'', \proxann{}.}
  \label{fig:teaser}
\end{figure}

\section{Introduction}
Suppose a researcher wants to study the impact of donations on politicians' speech~\cite{goel-2023-donor}.
For two decades, such questions have often been answered with the help of topic models or other text-clustering techniques \cite{jbg2017applicationstm}.
Here, the research team might interpret topic model estimates as representing \ul{healthcare} or \ul{taxation} categories, and associate each legislator with the topics they discuss.
Researchers could then measure the influence of a donation on the change in the legislators' topic mixture---showing that, e.g., money from a pharmaceutical company increases their focus on \ul{healthcare}.

The crucial supposition of such a ``text-as-data'' approach is that the interpreted categories are valid measurements of underlying concepts~\cite{Grimmer2013TextAD,ying2022topics,zhang-etal-2024-human}.
Adapting an example from \citeauthor{ying2022topics}, plausible interpretations of model estimates might yield either \ul{healthcare} or \ul{medical research}, which would carry ``very different substantive implications'' for a research area.
Facilitating the identification of valid categories is therefore a key concern in real-world settings, which falls under the framework of \emph{qualitative content analysis} \cite[\abr{qca},][]{Mayring2000}, a primary use case for topic models \cite{Grimmer2013TextAD,bakharia-etal-2016-interactive,hoyle-etal-2022-neural,li-etal-2024-improving}.

Taking the view that effective evaluations are those that approximate the real-world requirements of the use case \cite{liao-xiao-2023-rethinking}, it follows that topic model (and document clustering) evaluations should help encourage valid categories~\cite{ying2022topics}.
However, as we discuss in \cref{sec:background}, the evaluation strategies that are reasonable approximations for this use case are generally dependent on human-derived labels, rendering them hard to scale and reproduce.
Conversely, the most common unsupervised automated metrics, while fast to compute, are poor measures of topic quality \cite{hoyle-etal-2021-automated,doogan-buntine-2021-topic}.

This paper addresses these shortcomings by introducing both an application-grounded human evaluation protocol and a corresponding automated metric that can substitute for a human evaluator.
The protocol approximates the standard qualitative content analysis process, where categories are first derived from text data and subsequently applied to new items, \cref{fig:teaser}; our human study collects multiple annotations for dozens of topics, making it the largest of its kind.
Using both open-source and proprietary large language models (LLMs), we develop ``proxy annotators'' that complete the tasks comparably to an arbitrary human annotator; we call the method \proxann{}.
In addition, results from the human evaluation indicate that a classical model \cite[\lda{},][]{blei2003latent} performs at least as well, if not better, than its modern equivalents.\looseness=-1

\section{Background and Prior Work}\label{sec:background}

We outline necessary background regarding topic models, clustering methods and their evaluations.
We start with the goals of topic modeling, turn to standard automated evaluations, then outline use-oriented measures based on human input.

\subsection{Making sense of document collections}\label{sec:background:sensemaking}
The systematic categorization of text datasets is a common activity in many fields, particularly in the social sciences and humanities.
A common manual framework to help structure the recognition of categories in texts is \emph{qualitative content analysis} \cite[\abr{qca},][\emph{inter alia}]{Mayring2000,smith2000content,elo2008qualitative}.
Broadly, it consists of an inductive process whereby categories emerge from data, which are then consolidated into a final codeset.
These categories are then deductively assigned to new documents, supporting downstream analyses and understanding (e.g., characterizing the changing prevalence of categories over time).\footnote{
Practitioners in various communities have developed related families of methodologies with similar goals, such as \emph{grounded theory} \cite{glaser1967discovery} and \emph{reflexive thematic analysis} \cite{braun-clarke-2006-thematic})}

NLP offers techniques that are designed to support this process---and that are often conceived as analogous of such manual approaches \cite{bakharia-etal-2016-interactive,Baumer2017ComparingGT,hoyle-etal-2022-neural}.
These methods are typically unsupervised, and among the most prevalent are \emph{topic models} \cite{blei2003latent}.
A topic model is a generative model of documents, where each document is represented by an admixture of latent topics $\theta_d$, and each topic is in turn a distribution over words types $\beta_k$ (which a user can interpret as a category).
For example, when analyzing a corpus of U.S. legislation, suppose the most probable words for one topic include \texttt{doctor, medicine, health, patient} and a document with a high probability for that topic is the text of the \emph{Affordable Care Act}; together, they appear to convey a \ul{healthcare} category.

More recently, the improved representation capacity of sequence embeddings \cite[e.g., sentence transformers,][]{reimers-gurevych-2019-sentence} has led to their use in clustering (see \citealt{zhang-etal-2022-neural} for an overview).
As with topic modeling, a document is associated with one or more clusters (an equivalent to $\theta_d$); succinct labels (standing in for $\beta_k$) for clusters can be obtained with various word-selection methods or language-model summaries.

\subsection{Evaluating Categorizations}

\paragraph{Topic Coherence.} Topic model evaluation has primarily focused on the semantic \emph{coherence} of the most probable words in a topic---the capacity for a set of terms to ``enable human recognition of an identifiable category'' \citep{hoyle-etal-2021-automated}.
\citealt{BoydGraber2014CareAF} consider a topic's coherence to be a precondition for a useful model, and indeed, applied works often validate topics by presenting the top words \cite{ying2022topics}---which, in many cases, is the \emph{only} form of validation.
While \citealt{ying2022topics} attempt to standardize evaluations of topic-word coherence (building on \citealt{chang2009reading}), the reliance on crowdworkers renders them difficult and costly to scale.
As a result, methodological contributions---where easily-applied metrics can help guide model development---tend to use \emph{automated} proxies for coherence, like Normalized Pointwise Mutual Information \cite[NPMI,][]{lau-etal-2014-machine}.
Despite their ubiquity, automated coherence metrics fail to align closely with human judgments, exaggerating differences between topics \cite{hoyle-etal-2021-automated}.\footnote{\citealt{peng2024aligning} have investigated this relationship further, but with artificial topics not generated by a model.}
Newer automated metrics based on LLMs face similar issues, lacking a clear relationship to actual usage and human judgments of quality (details in \cref{sec:prior-work}).

\paragraph{Beyond Topic Coherence.} In contrast, our contribution closely matches standard qualitative analysis (\cref{sec:background:sensemaking}): developing and applying categories to text items.
Although coherent topic-words (or category labels) are important for interpretability, they are not sufficient to establish that model outputs are valid.
Categories are also assigned to individual text items, and those assignments should be ``meaningful, appropriate, and useful'' \cite{BoydGraber2014CareAF}.
Furthermore, the coherence of the topic-words may not agree with the perceived quality of the document-topic distribution \cite{bhatia-etal-2017-automatic}.
For topic models, \citealt{doogan-buntine-2021-topic} therefore argue that measuring the coherence of the top documents for each topic is necessary for a holistic model evaluation.\footnote{The same logic holds for document clustering, where the interpretation of a category relies on reading the documents assigned to it.}
Several prior efforts have situated model evaluation in the context of their use, but these works rely on on manual label assignments (either pre-existing or via interaction), limiting their broader utility (additional discussion of prior work in \cref{sec:prior-work}).

\section{\proxann{}}

\begin{table*}
  \centering
  \resizebox{\textwidth}{!}{
  \input{tables/topics_examples_table}

  }
\caption{\textcolor{customblue}{\textsc{ProxAnn-\gptFourO{}}} and human annotator-provided category labels for a sample of matched topics from each model (\emph{topic model words are in italics}) for \texttt{Wiki} (top row) and \texttt{Bills} (bottom row) datasets. Labels are consistent across humans and models.
}
\label{tab:category-labels}
\end{table*}

This section proposes a human evaluation protocol for topic models and document clustering methods.
The evaluation is oriented toward real-world use, emulating how practitioners develop categories from---and assign them to---text data in applied settings.
Alongside the human tasks, we also develop LLM prompts that adapt the human instructions, treating the LLM as a proxy annotator, \proxann{}.

In brief, a sample of documents and keywords for each topic or cluster are shown to an annotator to establish its semantic category (as in the first step in \citealt{ying2022topics}, who rely on experts to create labels); the annotator then reviews additional documents and labels them based on their relatedness to the category.
These \emph{category identification} and \emph{relevance judgment} steps follow that of qualitative content analysis, ``a manual process of inductive discovery of codesets via \emph{emergent coding}'' \citep{stemler2000overview}.
We also include a \emph{representativeness ranking} task as an additional evaluation signal, inspired by ``verbatim selection'' in qualitative settings \cite{corden2006using}.

As a whole, our proposal builds on the idea that coherence means ``calling out a latent concept in the mind of a reader'' \cite{hoyle-etal-2021-automated}.
By measuring the coherence of the documents within each topic or cluster, it provides a more holistic (and use-oriented) picture of a model's quality than past work.
It draws from the tasks in \citet{ying2022topics}; we adapt and combine their label assignment and validation steps, avoiding the reliance on curated expert labels.\footnote{However, our approach can also use expert labels, and is complementary to their work.}
Our protocol is also informed by interactive topic modeling for content analysis \cite{poursabzi-sangdeh-etal-2016-alto,li-etal-2024-improving,li2025largelanguagemodelsstruggle}, where topic model outputs help inform the creation and assignment of categories.

\begin{table}
\centering
\small
\begin{tabular}{llll}
\toprule
 &  & \multicolumn{2}{c}{Krippendorff's $\alpha$} \\
\cmidrule(lr){3-4}
 &  &  Fit Step & Rank Step \\
\midrule
 \multirow[c]{3}{*}{\texttt{Wiki}} & Mallet & 0.71 (0.10) & 0.74 (0.12) \\
 & \ctm{} & 0.55 (0.30) & 0.45 (0.11) \\
 & \bertopic{} & 0.57 (0.16) & 0.44 (0.20) \\
  \addlinespace[0.6ex]
\multirow[c]{3}{*}{\texttt{Bills}} & Mallet & 0.31 (0.27) & 0.49 (0.22) \\
 & \ctm{} & 0.37 (0.19) & 0.43 (0.26) \\
 & \bertopic{} & 0.32 (0.30) & 0.34 (0.17) \\
 \midrule
 \multicolumn{2}{c}{\emph{Label-Derived}} & 0.80 (0.13) & 0.86 (0.05) \\
\bottomrule
\end{tabular}
  \caption{
  Chance-corrected human--human inter-annotator agreement on the two annotation tasks (Krippendorff's $\alpha$), averaged over eight topics per model (standard deviation in parentheses). \emph{Label-Derived} are six clusters derived from ground-truth \texttt{Wiki} labels used in a pilot study, serving as a high-coherence reference. High agreement on the reference indicates the tasks are well-specified. (Topics have $\geq3$ annotators; variation due to filtering out low-quality annotators).
}
\label{tab:topic-agreement}
\end{table}

\subsection{Evaluation Protocol}\label{subsec:protocol}

We describe the steps for the human evaluation protocol and LLM-proxy, \proxann{}, in parallel.
Appendices contain instructions, user interface screenshots (App. \ref{app:interface}), and model prompts (App. \ref{app:prompting}).

\paragraph{Setup.}\label{subsec:definitions} First, we outline the model outputs required for the evaluation (recall that we are attempting to emulate content analysis, \cref{fig:teaser}).
Throughout, we remain as agnostic as possible to the method that produces these outputs; the evaluation is appropriate for both topic models and other text clustering techniques.

Suppose that there are $K$ topics or clusters and $|\mathcal{D}|$ documents, with each document containing $|W_d|$ word types (total vocabulary size $|W|$).
Each document $d \in \mathcal{D}$ has an estimated score indicating its semantic relationship to the $k$th topic or cluster, $\theta_{dk}$.
For topic models, this is the estimated posterior probability for the $k^\text{th}$ topic.
Different clustering methods can produce this value in different ways; e.g., for $K$-means, a standard estimate is the similarity between the document embedding and the cluster centroid.
We place estimates into a matrix $\bm{\Theta} \in \mathbb{R}^{N \times K}$, and each column of the matrix sorted to produce a ranked list of the most likely documents for each topic or cluster, $\bm{\theta}^{(r)}_k$.

Topics and clusters are also associated with ranked word types $\bm{\beta}^{(r)}_k$.
For topic models, these are the sorted rows of the topic-word distributions $\mathbf{B} \in \mathbb{R}^{K \times |W|}$; for clustering, it is possible to extract top words for a cluster via tf-idf~\cite{sia-etal-2020-tired}.\footnote{Ranked word types are not strictly necessary for the evaluation, but their usage as a topic summary is widespread.}\looseness=-1

The final representations shown to users consist of a sample of $n_{d}$ highly-ranked \textbf{exemplar documents} from $\bm{\theta}^{(r)}_k$ and the most probable $n_w$ \textbf{keywords} from $\bm{\beta}^{(r)}_k$.
To balance informativeness with annotator burden, we set the number of documents $n_{\text{ex}}$ to seven and the number of words $n_w$ to 15.\footnote{See \citealt{lau-baldwin-2016-sensitivity} for a discussion of the relationship between $n_w$ and perceived coherence.}
Exemplar documents are a stratified sample over $\bm{\theta}_k$ (details in App. \ref{app:exemplar-selection}).

\paragraph{\ul{Label Step}: Category identification.} %
After viewing instructions and completing a training exercise (App. \ref{app:interface}), each annotator reviews the exemplar documents and keywords for a single topic. They then construct a free-text \textbf{label} that best describes the category they have observed.\footnote{Per \citet{chang2009reading}, documents are truncated to improve reading times. We limit them to 1000 characters.} Continuing the earlier U.S. \ul{healthcare} example, users might also view the text of the \emph{National Organ Transplant Act} and the \emph{Rare Diseases Act}.

The LLM is prompted with condensed instructions and the same exemplars and keywords, also producing a label for the category.

\paragraph{\ul{Fit Step}: Relevance Judgment.} An additional sample of seven \textbf{evaluation documents}, evenly stratified over $\bm{\theta}^{(r)}_k$, is shown in random order.\footnote{Generally, we assume a strict total ordering over evaluation documents; nonstrict orders, as in the case of binary assignments $\theta_{dk} \in \{0, 1\}$, can work but require some alterations to our metrics.}
For one document at a time, annotators answer the extent to which the document fits their inferred category (on a scale from ``1 -- No, it doesn't fit'' to ``5 -- Yes, it fits''), producing a set of \textbf{fit scores} for annotator $i$, $\bm{s}^{(i)}_k$.
As a control, one document with near-zero probability for the topic is always shown.
Here, an annotator might assign the \emph{Coronavirus Preparedness and Response Act} a ``5'' and the \emph{Federal Meat Inspection Act} a ``3''.

For the LLM prompt, we take the probability-weighted mean over tokens in the scale to obtain relevance judgments, per \citet{wang2025improvingllmasajudgeinferencejudgment}: $\sum_{s\in\{1\ldots5\}} s\cdot p_{LM}(\texttt{s}\mid \texttt{instruction}, \texttt{doc}_i)$.

\paragraph{\ul{Rank Step}: Representativeness ranking.} Last, annotators rank the evaluation documents by how representative they are for that category, $\bm{r}^{(i)}_k$.\footnote{We include a ``distractor'' document---an Amazon review for kitchen sponges---to filter out poor quality annotations.}

Given the complexity of the task, a direct translation to an LLM prompt is not practical.
Instead, we modify the question to include two evaluation documents at a time, leading to $7 \choose 2$ prompts.
The LLM thus produces a set of pairwise ranks per prompt, which we use to infer real-valued ``relatedness'' scores for each document with a \citeauthor{bradley_rank_1952} model  (further details in App. \ref{app:prompting}).

\section{Experimental Setup}
\label{sec:experiments}
We describe the experimental setup: the choices of datasets, models, annotators, and metrics.

\subsection{Datasets}
 We use two English datasets that are standard in topic modeling evaluations \cite[e.g.,][]{pham-etal-2024-topicgpt,lam-etal-2024-lloom,li-etal-2024-improving,zhong2024explaining}: \texttt{Wiki}~\cite{merityPointerSentinelMixture2017}, a general audience-corpus consisting of 14,000 ``good'' Wikipedia\footnote{\url{https://en.wikipedia.org/wiki/Wikipedia:Good_article_criteria}} articles; and \texttt{Bills}~\cite{adler2018congressional}, a more specialized domain-specific dataset comprising 32,000 legislative summaries from the 110th--114th U.S. Congresses. We use the preprocessed version of these datasets from \citealt{hoyle-etal-2022-neural}, in its 15,000-term vocabulary form.

\begin{figure*}[ht]
      \centering
      \hspace*{-1.35cm}
    \begin{subfigure}{0.49\textwidth}
          \centering
        \includegraphics{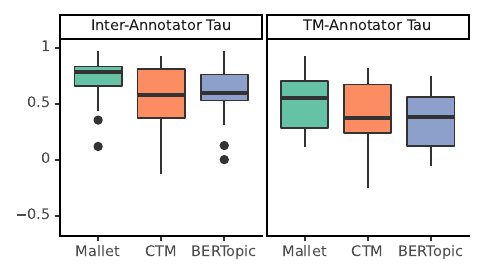}
        \caption{Relevance Judgements (\emph{Fit Step})}\label{fig:topic-human-scores:fit}
    \end{subfigure}
    \begin{subfigure}{0.49\textwidth}
          \centering
        \includegraphics{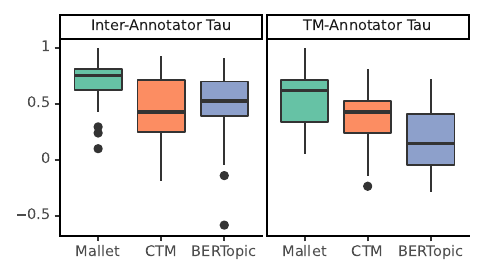}
        \caption{Representativeness Ranking (\emph{Rank Step})}\label{fig:topic-human-scores:rank}
    \end{subfigure}
  \caption{
  Annotators review the top documents and words from a single topic and infer a category (Label Step), then assign scores to additional documents based on their relationship to the category (Fit and Rank Steps). These scores are correlated with each other (Inter--Annotator Kendall's $\tau$) and with the model's document-topic estimates ($\bm{\theta}_k$; TM--annotator $\tau$). There are eight topics per model; boxplots report variation in $\tau$ over each topic-annotator tuple.
  }
  \label{fig:topic-human-scores}
\end{figure*}

\subsection{Models}
\paragraph{Topic Models} Topic models can be broadly categorized into \emph{classical} Bayesian methods, which use Gibbs sampling or variational inference to infer posteriors over the latent topic-word ($\bm{B}$) and document-topic ($\bm{\Theta}$) distributions, and \emph{neural} topic models, often estimated with variational auto-encoders~\cite{kingma2013auto}.
Clustering techniques can also approximate topic models; in a typical setup \cite[e.g.,][]{zhang-etal-2022-neural}, $K$-means is applied to sentence embeddings \cite[\abr{sbert},][]{reimers-gurevych-2019-sentence} of the documents.\footnote{Recently, LLM-based topic models \cite{pham-etal-2024-topicgpt,lam-etal-2024-lloom} offer more ``human-readable'' topic descriptions, but lack the document-topic and word-topic distributions that other methods provide or approximate. Given these differences, we leave an evaluation to future work.}

We evaluate one model from each class: \lda{}~\cite{blei2003latent} using the \mallet{} implementation (hereafter referred to as \textbf{\mallet{}}), \textbf{\ctm{}}~\cite{bianchi-etal-2021-pre}, and \textbf{\bertopic{}}~\cite{grootendorst2022bertopic}. We reuse the 50-topic \mallet{} and \ctm{} models from \citealt{hoyle-etal-2022-neural} and train \bertopic{} under the same experimental setup using default hyperparameters (details in App. \ref{app:bertopic_training_details}).
In a pilot study, we also evaluate a semi-synthetic upper bound model derived from ground-truth \texttt{Wiki} labels (App. \ref{app:pilot}).

\paragraph{\proxann{} LLMs} As LLM annotators, we use OpenAI's \gptFourO{} (\texttt{gpt-4o-mini-2024-07-18}), \llamaThreeOneEightB, \llamaThreeThreeSeventyB{} \cite{dubey2024llama3herdmodels}, \qwenTwoFiveSeventyTwoB, and Qwen 3 in both 8B and 32B variants \cite{qwen2025qwen25technicalreport,yang2025qwen3technicalreport}. For the label generation step, we set the temperature to 1.0. Then, conditioned on that label, the Fit and Rank steps have temperature 0.\footnote{Documents exceeding 100 tokens are truncated, extending to the end of the sentence to avoid incomplete cuts.}
We then \emph{re-sample} the chain of steps five times and take the mean responses (just as we average over human annotators).
More details can be found in App. \ref{app:prompting}.

\subsection{Collecting Human Annotations}

A comprehensive human evaluation of all topics would be cost-prohibitive, so we randomly sample eight of the fifty topics for the \texttt{Wiki} and \texttt{Bills} data on each of the three models.
We recruit at least four annotators per topic through Prolific. 
Low-quality respondents are filtered out with attention checks.\footnote{\url{prolific.com}, further recruitment details in App. \ref{app:annotators}. While using more annotators per topic would provide more robust estimates of model performance, \citet{ying2022topics} use two per topic in a similar setup; the statistical test of annotator--LLM substitutability (\S\ref{subsec:metrics}) requires only three. Agreement is also high for a synthetic upper-bound (\cref{tab:topic-agreement}).}

Model-to-model results on a subset of topics may not be comparable; when sampling, we first pick a random topic from one model, and choose the topics from the remaining models with the smallest word-mover's distance \cite[computed using word embeddings of the topic words,][]{kusner2015word, flamary2021pot}.

\paragraph{Initial Task Validation}There are two potential sources of disagreement in the resulting annotations: either the topics can be incoherent, leading to inconsistent category labels between annotators, or the tasks could be poorly defined.
As an initial validation step, we run a pilot study with six clusters derived from ground-truth \texttt{Wiki} labels to serve as a rough upper bound (comparing them with six \ctm{} and \mallet{} topics as a reference).
Specifically, we create clusters by assigning all documents to their labeled category (e.g., \textsc{Media and drama}), then rank the documents within a cluster based on their cosine similarity to the centroid (computed with \abr{sbert} \cite{reimers-gurevych-2019-sentence}; additional details and results in App. \ref{app:pilot}).

This approach reduces noise introduced by low-quality topics: because annotators review documents that belong to the same coherent category, their inferred conceptualizations should be fairly crisp.
If the tasks are well-specified, annotations on the tasks should be consistent.

\begin{table}
\centering
\small
\resizebox{\columnwidth}{!}{%
\input{tables/stat_test_table}
}
\caption{
  Advantage probabilities from the alternative annotator test; the probability that \proxann{} is ``as good as or better than a randomly chosen human annotator'' \cite{calderon2025alternativeannotatortestllmasajudge}. Document-level scores consider annotations by document; Topic-level over all documents evaluated in the topic. $*$ indicates that win rates over humans are above 0.5, as determined by a one-sided t-test (over 10 resamples of combined annotators). $\dagger$ is the equivalent for Wilcoxon signed-rank.
}\label{tab:alt_test}
\end{table}

\subsection{Metrics}\label{subsec:metrics}
We examine four aspects of our approach: the sensibility of the human evaluation protocol; using the protocol to evaluate topic models and clustering; comparing human annotations with the LLM proxy; and using metrics based on the LLM proxies to score topics and clusters.

\paragraph{Human--human agreement on the tasks.} Following standards from the content analysis literature, we use \citeauthor{krippendorff-2019-content}~'s $\alpha$ to assess the chance-corrected agreement across human annotators for the Fit and Rank steps (with ordinal weights).\footnote{Although it seems natural to use these metrics for model comparisons---higher agreement indicating better topics---there are complications arising from skewed distributions and respondents annotating one topic at a time, App. \ref{app:agreement}.}
For easier comparison with the topic model--human metrics (next section), we also compute annotator-to-annotator correlations between each annotator's relevance fit scores or ranks and the averaged fits (ranks) of all other annotators.

\paragraph{Human evaluation of topics and clusters.} Per \cref{subsec:protocol}, models estimate real-valued scores $\theta_{dk}$ that (should) correspond to the relevance that document $d$ has for category $k$.
In the Fit and Rank steps, annotators assess the relevance of seven documents over a stratified set of these scores for a topic $k$, $\bm{\theta}^{\text{eval}}_{k}$ (all annotators review the same documents; see App. \ref{app:exemplar-selection} for $\bm{\theta}^{\text{eval}}$ sampling details).

As a measure of model quality, we report the correlation coefficients for Kendall's $\tau$ \cite{Kendall1938} to measure both annotator-model and inter-annotator relationships.
The annotator-model correlations are between the estimated probabilities per document $\bm{\theta}^{\text{eval}}_{k}$ with either the human relevance scores ($\bm{s}^{(i)}_k$, Fit Step) or their ranks ($\bm{r}^{(i)}_k$, Rank Step), for annotator $i$.
We contextualize these against the inter-human-annotator $\tau$ (see above).

\paragraph{\proxann{}--human agreement.} For the LLM to serve as a proxy, it should ideally be indistinguishable from a human annotator.
This idea is operationalized by the Alternative Annotator Test \cite[alt-test,][]{calderon2025alternativeannotatortestllmasajudge}. %
For each annotated instance $d_i$, the alt-test computes two leave-one-out similarity metrics: the similarity between between annotator $j$'s responses and the responses of all \emph{other} annotators, $s_{h:h}(d_i)$, and the similarity between the LLM's response and those of all other annotators, $s_{lm:h}(d_i)$.
The result is a set of binary outcomes $\mathbb{I}\left[s_{h:h}(d_i) < s_{lm:h}(d_i)\right]$, and a one-sided t-test (Wilcoxon signed-rank for small $n$) determines whether the LLM wins significantly more often (subject to a slack term $\epsilon$ we set to $0.1$, per their suggestion for crowdworkers).
It computes $\omega$, the (multiple-comparison-adjusted) win rate of LLMs over annotators (over all annotators $j$), and the \emph{advantage probability} that an LLM is as (or more) reliable than a human,  $\rho$.

We apply the alt-test to annotations on individual documents as well as on the entire topic, using the root mean squared error as the similarity metric.
Our annotators independently review only seven documents from a single (model, topic, dataset) tuple, which is insufficient for the test.
Following \citeauthor{calderon2025alternativeannotatortestllmasajudge}'s recommendation, we combine annotations to create pseudo-annotators. We combine over all topics within a dataset, such that the ``annotator'' observes $n_{\text{models}}\cdot n_{\text{topics}} \cdot n_{\text{docs}}=3\times8\times7$ items (we bootstrap over ten random permutations, computing $\omega$ over the full set; variance of $\rho$ is small and not reported. Results from combining over topics per model in App. \ref{app:alt_test_model_combs}.)

\paragraph{\proxann{} as an automated evaluator.} A common use for automated coherence metrics, like NPMI \cite{lau-etal-2014-machine}, is the ranking of topics---and the averaging of topics within each model to rank models.
Indeed, NPMI is the dominant metric used in the literature to compare proposed models against baselines \cite{hoyle-etal-2021-automated}.

We compare the human evaluations of topics and clusters to metrics based on \proxann{}.
Define evaluation metrics per the above descriptions: 
\begin{align}
    \fittauHTM{}(k) &:=\tau \left( \bm{s}^{(h)}_k,\bm{\theta}^{\text{eval}}_{k} \right)\\
    \fittauLMTM{}(k)&:=\tau \left( \bm{s}^{(lm)}_k,\bm{\theta}^{\text{eval}}_{k} \right).
\end{align}
These are the correlations for topic $k$ between the Fit step responses (from either humans or \proxann{}, $\bm{s}_k$) and (b) the estimated document scores from the topic model. Define $\textsc{rank–}\tau(k)$ analogously for the Rank-step responses.
For each topic and task, there is a ``ground-truth'' evaluation metric  and a ``proxy'' metric --- $\fittauHTM{}(k)$ or $\ranktauHTM{}(k)$ ---and a ``proxy'' metric, $\fittauLMTM{}(k)$ or $\ranktauLMTM{}(k)$.
A second Kendall's $\tau$ over these metrics for all $k$ measures the extent to which \proxann{}'s rankings over topics agrees with that of the average human.\looseness=-1

\section{Results} 
\begin{table}[!t]
\centering
\label{tab:topline_table}
\resizebox{\columnwidth}{!}{%

\input{tables/topline_table}
}
\caption{Relationship between automated and human topic rankings. Cells show Kendall's $\tau$ between metrics: $\textsc{Fit/}\ranktauHTM{}$ correlates human scores to document–topic probabilities ($\bm{\theta}_k$); $\textsc{Fit/}\ranktauLMTM{}$ correlates \proxann{} to $\bm{\theta}_k$. Values are bootstrapped means and standard deviations (resampling over topics). The \emph{Human} row reflects leave-one-out inter-annotator correlations, serving as a reference. While larger Qwen models achieve the strongest correlations, \gptFourO{} is middling. NPMI is not correlated with human metrics.}

\end{table}

We discuss results in the same order they were presented above. In tables and figures, \textbf{Fit} refers to responses to  the relevance judgments of evaluation documents and \textbf{Rank} to responses to representative rankings of the documents.

\subsection{Human--Human Agreement}
Generally, annotators respond consistently, providing qualitatively sensible labels to the topics (\cref{tab:category-labels}).
Average agreement per topic (Krippendorff's $\alpha$) is reasonably strong overall, particularly for the ranking tasks on the \texttt{Wiki} data (\cref{tab:topic-agreement}).
We emphasize that \emph{low} agreement is likely indicative of a poor model, rather than a misspecified task: the agreement metrics for the synthetic \emph{label-derived} clusters are very strong ($\alpha\geq0.8$ on both tasks).
Overall, \mallet{} tends to have higher agreement; however, variance over topics is somewhat high, and we caution against using $\alpha$ for model comparisons.
Together, these results point to the viability of our evaluation protocol, implying that the demands of the tasks are intelligible and reproducible.

\begin{figure*}[ht]
  \centering
  \includegraphics[width=\textwidth]{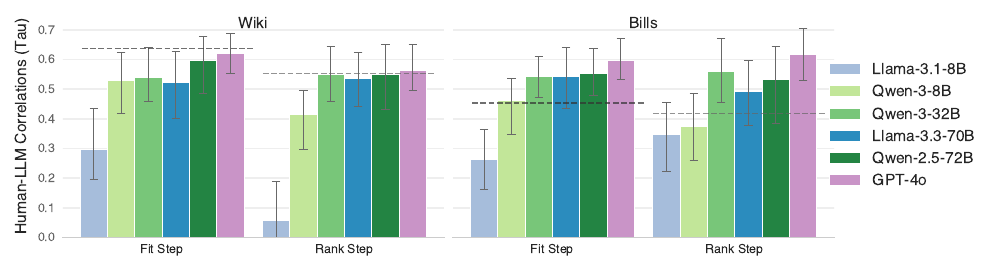}
  \caption{
    Correlations between \proxann{} and human annotations (Kendall's $\tau$) for the relevance judgment (Fit Step) and representativeness ranking (Rank Step) tasks, averaged over topics (pooled over all three models). While \gptFourO{} has the best overall correlations and relatively low variance, the Qwen family is a reasonable substitute, even at smaller sizes. Dashed horizontal lines are the average leave-one-out human--human correlation (per \cref{fig:topic-human-scores,fig:topic-human-scores-all-bills}). Error bars represent 95\% bootstrapped CIs, resampling over topics.
    }
  \label{fig:human-llm-barplot}
\end{figure*}
\subsection{Human Evaluations of Topics}

Our protocol creates consistent and sensible results.
There is generally a positive correlation between the estimated document-topic probabilities ($\bm{\theta}_k$) and human judgments on the \texttt{Wiki} data (\cref{fig:topic-human-scores}, \texttt{Bills} data in \cref{fig:topic-human-scores-all-bills} in the appendix).
Comparing the first two plots (human--human) to the second two (human--model), annotator agreement with other annotators is generally higher than annotator agreement with the model.
Both the inter-annotator and model--annotator scores show a consistent ranking over models: \mallet{} fares better than \ctm{}, and \ctm{} better than \bertopic{}---in fact, several topics have negative correlations for \bertopic{}.
In App. \ref{app:additional-metrics}, we report on two additional metrics, NDCG and binarized agreement.

These results support the idea that \mallet{}, at 20 years old, remains an effective tool for automated content analysis (echoing \citealt{hoyle-etal-2022-neural}).

\subsection{Is \proxann{} a good proxy?}

Generally, \proxann{} is a reasonable proxy across both steps and datasets, although there is variation among the underlying LLMs.
The correlations between LLM and human responses are generally around $\tau=0.5$ or greater for the largest ($\geq32\text{B}$) models (\cref{fig:human-llm-barplot}); \qwenThreeEightB{} is competitive in a few cases, but \llamaThreeOneEightB{} is generally poor.
In many cases, the best models meet or exceed the average leave-one-out human-to-human correlations.

These results are largely corroborated by the alt-test (\cref{tab:alt_test}).
When considering annotations at the document-level, \proxann{} (for larger models) is a suitable substitute for human annotators: advantage probabilities $\rho$ are generally over 0.5 for the stronger models, and have significantly higher agreement rates with humans than humans do with each other.
The picture is a little less rosy for the topic-level annotations, where agreement is computed as the $\tau$.
Here, the tests are lower-powered (as the number of instances has been reduced by a factor of $n_\text{docs}=7$), making statistical wins less probable.
Second, the high human agreements for the relevance judgment (fit) tasks on the \texttt{Wiki} data make it harder for an LLM to perform as well. 
In addition, it may be that the LLM provides overly-specific topics on Wikipedia due to greater domain ``knowledge'' (further discussion in \cref{sec:agreement_eval}).%

\subsection{Ranking Topics and Models}
We now measure whether metrics derived from \proxann{} rank topics and models similarly to humans.
Generally, no model is dominant, with almost all correlations less than $\tau=0.5$.
\qwenTwoFiveSeventyTwoB{} performs reasonably well on the \texttt{Wiki} data, but performance on \texttt{Bills} is generally low.
There, low correlations may be attributed to (a) a more specialized dataset requiring additional background knowledge and (b) having tuned prompts on pilot annotations from the \texttt{Wiki} data.

While not very high, these values are comparable to leaving out one \emph{human} annotator and computing their agreement with the average of the other humans (e.g., the mean \texttt{Wiki} $\tau$ for the rank task is 0.33).
We also report results with \emph{binarized} $\bm{\theta}$, corresponding to hard assignments, which tend to show better correlations (App. \ref{app:model_ranks}).
Meanwhile, the standard automated metric, NPMI, fails to capture the human judgments.

Together, these results indicate that there is some capacity for \proxann{} to accurately rank topics at least as well as an arbitrary annotator.
Last, when aggregating at the \emph{model}-level (i.e., over \ctm{}, \bertopic{}, \mallet{}) model rankings align for \texttt{Wiki} and are generally close for \texttt{Bills} (App. \ref{app:model_ranks}).

\subsection{A Qualitative View of Agreement}\label{sec:agreement_eval}
Last, we examine how annotator agreement---and agreement between annotators and the topic model or LLM---reflects topic quality (\cref{app:agreement}). %

Topics with low human-to-human agreement tend to be too broad or multi-themed, leading to disagreement in both categorical labels and document fits across human annotators (\cref{tab:disagreement_human_tm_evaluation} in appendix). In contrast, low human-to-topic model agreement (conditioned on high human--human agreement) often reveals model-specific limitations: \bertopic{} may underassign relevant documents due to its hard clustering approximation, while \mallet{} may fully assign a document to a single topic (\(\theta_d = 1\)) simply because no better alternative is available. This pattern highlights that under high human--human agreement, low human--topic model agreement is more likely to indicate model failure than annotation ambiguity.

We also analyze two case studies of high-human agreement topics where LLM judgments nonetheless diverge (\cref{tab:example_human_llm_disagreement1,tab:example_human_llm_disagreement2} in appendix). 
While the human-to-LLM agreement yields moderate Kendall's $\tau$ values (0.48 and 0.58), qualitative inspection shows that actual differences in fit judgments are often small. Table~\ref{tab:example_human_llm_disagreement1} shows near-perfect alignment between human and LLM fit scores, despite the lower $\tau$. In Table~\ref{tab:example_human_llm_disagreement2}, where $\tau$ is slightly higher, fit scores also agree closely, though minor differences emerge from how the topic is interpreted. Together, these examples raise the question of what constitutes a ``low'' or ``bad'' $\tau$ in this context; values below 0.5 may still reflect reasonable alignment. Interpreting $\tau$ in isolation may be misleading, and thresholds for ``good'' agreement should be grounded in qualitative examples.

\section{Prior Work}\label{sec:prior-work}
\paragraph{Use-oriented evaluations} \citealt{poursabzi-sangdeh-etal-2016-alto} and \citealt{li-etal-2024-improving} invoke topic models' usage in content analysis settings to inform new interactive methods, which are evaluated by measuring the alignment between method outputs and ground-truth labels. %
In a different use-inspired approach closer to our protocol, \citealt{ying2022topics} propose crowdworker ``label validation'' tasks, designed to assess the quality of individual document-topic distributions using already-identified expert labels.
Furthermore, the tasks require a curated set of labels covering \emph{all} relevant topics, whereas our setup can assess topics independently.
Although the above evaluations are better aligned with real-world use than topic coherence, they rely on some form of manual labeling, and are difficult to scale (\citealt{ying2022topics} only evaluate on one dataset).

\paragraph{LLM-based evaluations.} Metrics based on LLMs have become increasingly common in the NLP literature, notably in machine translation and human preference modeling \cite{zheng-etal-2023-judging}.
Within topic modeling, past efforts construct prompts designed to replicate human annotation tasks.
Both \citealt{stammbach-etal-2023-revisiting} and \citealt{rahimi-etal-2024-contextualized} prompt LLMs to emulate the word intrusion and rating tasks from \citealt{chang2009reading}, but these tasks assess only the top topic-words, an incomplete view of model outputs.
In addition, the correlations with human judgments are also mixed, with standard automated coherence metrics performing better in some cases.\footnote{\citealt{stammbach-etal-2023-revisiting} also propose an alternative document-labeling metric, but it is used for selecting an optimal number of topics, rather than measuring overall quality.}
In \citealt{yang2024llmreadingtealeaves}, a topic model and an LLM separately produce keywords to label documents: if the keywords tend to align, then this indicates a good model.
Although LLM keywords align well with human-generated ones for one of two datasets, the metric does not assess the overall cohesiveness of topics, and so the connection between this task and real-world use is unclear.\looseness=-1

\section{Conclusion} The quality of models is determined their ability to meet real-world needs \cite{liao-xiao-2023-rethinking}.
This work aims to meet those needs by designing a human evaluation protocol and corresponding automated approximation, \proxann{} that together reflect practitioners' real-world usage of topic models and clustering methods.
We anticipate that both the collected human evaluation data and automated approach will inspire future work in improving models, metrics, and downstream usage.

There are several promising directions in these areas: the development of specialized models for automated topic and cluster annotation, rather than generalized LLMs; extending our approach to non-English languages; incorporating annotations from experts who have specific information-seeking needs.
To support adoption and further experimentation, we provide both a demo interface\footnote{See link at \url{https://github.com/ahoho/proxann}} and a local deployment option for computing \proxann{} metrics on new model outputs.

\section*{Limitations} A primary limitation of our LLM-proxy is that it is a substitute for a \emph{single} human annotator.
However, a strong indicator of a poor cluster or topic is disagreement among \emph{multiple} annotators.
In future work, we intend to model disagreement directly, e.g., following recent approaches for finetuning reward models in the presence of human disagreement \cite{zhang2024divergingpreferencesannotatorsdisagree}, or earlier work on Bayesian models of annotation \cite{paun-etal-2018-comparing}.
Addressing this issue could also help solve another limitation: LLMs are more costly to deploy than previous automated metrics, but a model finetuned for this task could be smaller. 

Another shortcoming of our approach is the use of crowdworkers.
Although we use several mechanisms to ensure high-quality annotators (training questions, multiple comprehension and attention checks, requiring a bachelor's degree or higher, bonuses for good responses), the annotators are not experts pursuing a research question.
That said, we believe our use of multiple annotators per topic, along with the filtering described, ensures annotations of reasonably high quality (as seen by the consistent labels and annotations).\footnote{A reviewer also suggested combining LLM and human annotations, per \citet{he2024crowdsourced}.}
Exploring the role of expertise in topic model evaluation is an important direction for future research---in particular, studying the relationship between expert annotations and those from language models (as well as how they differ from crowdsourced annotations).

A final limitation is our exclusive use of English-language datasets. 
While we do not have access to the exact pretraining mixture for the LLMs, it is reasonable to assume that English data is a dominant component, in addition to being heavily favored in evaluation tasks.
We therefore do not expect our findings to generalize directly to non-English settings.

\section*{Acknowledgments}

Many thanks are owed to Rupak Sarkar for discussion and insight during initial planning stages of this work. Kris Miler, Aaron Schein, and Michelle Mazurek provided comments on the human evaluation protocol, which appeared in Alexander's dissertation. 
We also gratefully acknowledge Saúl Blanco Fortes for his valuable help with system configuration and the setup of the ProxAnn test frontend. The work of Lorena Calvo-Bartolomé has been partially supported by Grant PID2023-146684NB-I00, funded by MICIU/AEI/10.13039/501100011033 and by ERDF/UE.
This work was also supported in part by the U.S. National Science Foundation awards 2124270 (Resnik) and 2229885 (Boyd-Graber).
Any opinions, findings, conclusions, or recommendations expressed in this material are those of the author and do not necessarily reflect the views of the National Science Foundation.

\bibliography{custom}

\appendix

\section{Exemplar Document Selection}\label{app:exemplar-selection}
When constructing the exemplar documents, \citet{doogan-buntine-2021-topic} note that only showing the documents at the head of the distribution can lead to an overly-specific view of the topic (e.g., ``banning AR-15s'' vs. ``gun control'').
We mitigate this issue by instead sampling documents with a $\theta_{dk}$ greater than a threshold $t_k$.
To set $t_k$, we find the point with maximum curvature using an ``elbow''-detection algorithm \cite{Satopaa2011FindingA}.
Then, we sample from the set $\{ d : \theta_{dk} > t_k\}$, where the probability of a sample is proportional to $\theta_{dk}$.
\Cref{fig:theta-distributions} (in the appendix) shows the distributions of $\bm{\theta}^{(r)}_k$ for the 1,000 documents with the largest values over six topics for the two topic models we use (see \cref{sec:experiments}).

In \cref{fig:theta-distributions}, we visualize these distributions for \ctm{} and \mallet{} for the pilot topics alongside the detected threshold.
Documents above this threshold are sampled (proportional to $\theta_{dk}$ to produce the exemplar documents .

\begin{figure}[ht]
    \centering
    \begin{subfigure}[b]{0.49\textwidth}
        \centering
        \includegraphics{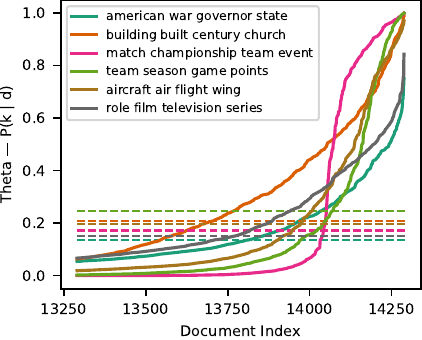}
        \caption{\mallet{}}
    \end{subfigure}
    \hfill
    \begin{subfigure}[b]{0.49\textwidth}
        \centering
        \includegraphics{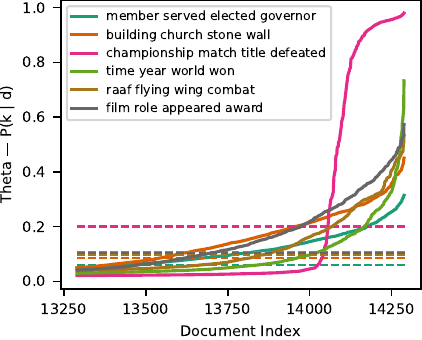}
        \caption{\ctm{}}
    \end{subfigure}
    \caption{
        Distribution of the top 1,000 theta values across six topics for two models.
        Topics have been aligned between models based on the word-mover's distance \cite{kusner2015word}.
        Dashed lines correspond to automatically determined ``elbows'' that threshold the $\bm{\theta}_k$ to produce representative documents.
        Some topics, like the \ul{championship} topic (in pink), have a sparser distribution and steep dropoff in values; others, like the \ul{building} topic (orange), have a more gradual decline in value.
    }
    \label{fig:theta-distributions}
\end{figure}

\section{Annotator Recruitment}\label{app:annotators}
Annotators must be fluent in English and have a college degree or higher.
Given the western-centrism of the English \texttt{Wiki} data respondents must be located in the U.S., Canada, Ireland, or U.K.; for the U.S.-centric \texttt{Bills} data, we exclude those outside North America.
We recruit at least 4 annotators per topic using Prolific. 
Demographic information is not made available to us, and we retain no identifying information.
Annotators were presented with information about the nature of the task and asked to provide consent before participation.
We set pay at a 15 USD per hour equivalent (\texttt{Wiki} completion time was estimated at 15 minutes, paying 3.75 USD per survey; \texttt{Bills} was updated to 4.25 for 17 minutes).
To encourage careful responses, we instruct annotators to ``give the answers you think most other people would agree with'', awarding a 1.50 USD bonus to those who have over 0.75 correlation with the average ranking of the other annotators for that topic.
Annotators who fail attention checks are not awarded a bonus and are excluded from the data.
An ethics review board deemed this study to not be human subjects research, and therefore exempt from review.

\section{Additional Results from the Human Study}\label{app:additional-metrics}
\begin{figure*}[ht]
  \centering
  \includegraphics[width=\textwidth]{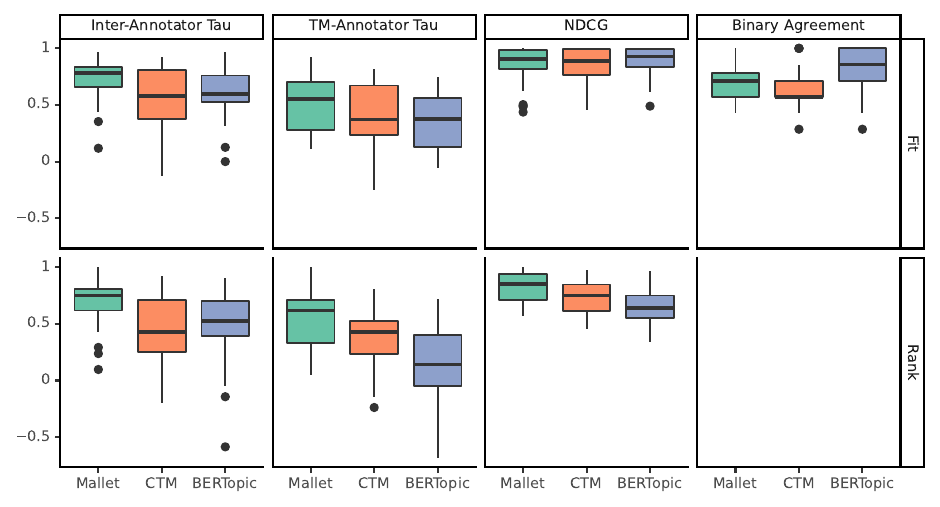}
  \caption{
    Metrics quantifying the relationship between human annotations and estimated document-topic probabilities ($\bm{\theta}_k$) for the three topic models on all eight \texttt{Wiki} topics. From left-to-right, the metrics are inter-annotator Kendall's $\tau$, model-annotator $\tau$, relevance agreement, and NDCG. The top row of figures reports relationships with human relevance judgments (on a 1-5 scale, Fit Step in the protocol), and the bottom row relationships with their document rankings (Rank Step). Boxplots report variation over topic-annotator pairs (binary agreement does not apply to the rank task).
  }
  \label{fig:topic-human-scores-all-wiki}
\end{figure*}

\begin{figure*}[ht]
  \centering
  \includegraphics[width=\textwidth]{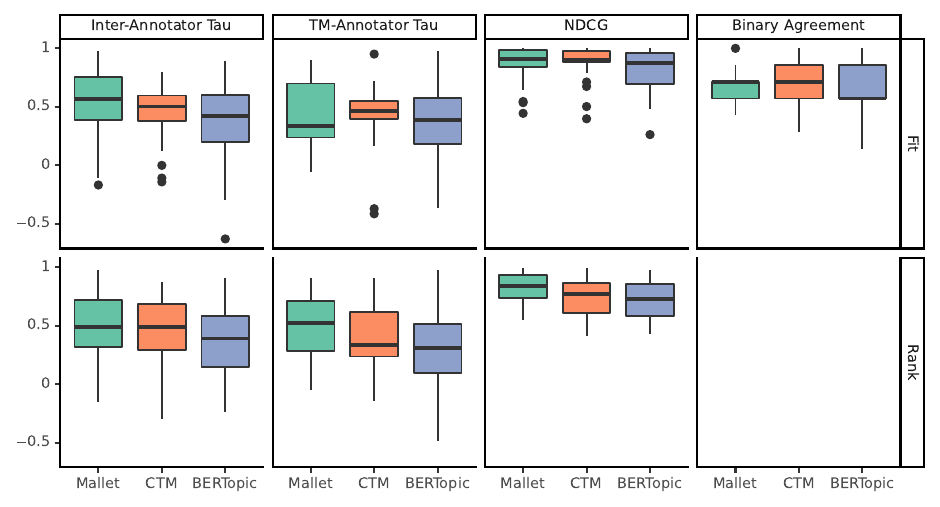}
  \caption{
    Metrics quantifying the relationship between human relevance judgments and estimated document-topic probabilities ($\bm{\theta}_k$) for three models on all eight \texttt{Bills} topics. See \cref{fig:topic-human-scores-all-wiki} for additional details.
  }
  \label{fig:topic-human-scores-all-bills}
\end{figure*}

\begin{figure*}[ht]
    \centering
    \begin{subfigure}[b]{\textwidth}
        \centering
        \includegraphics{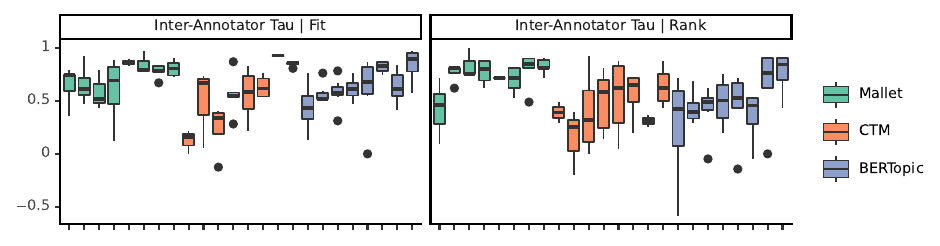}
        \caption{Wiki}
    \end{subfigure}
    \hfill
    \begin{subfigure}[b]{\textwidth}
        \centering
        \includegraphics{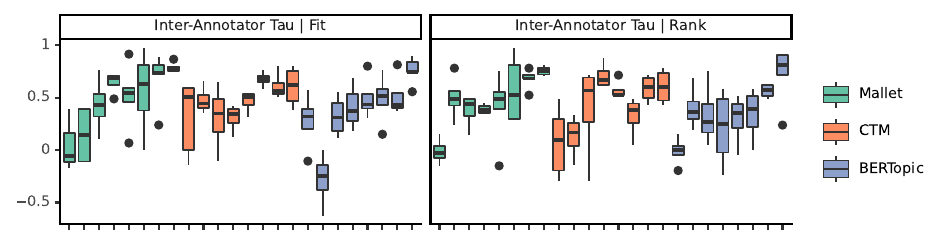}
        \caption{Bills}
    \end{subfigure}
    \caption{
        Distributions of leave-one-out inter-annotator correlations (Kendall's $\tau$) over all topics. Boxplots report variation over the annotators.
    }
    \label{fig:boxplots-by-topic}
\end{figure*}

In this section, we report on additional measures for the \textbf{human evaluations of topics} (\cref{subsec:metrics}).

We use Normalized Discounted Cumulative Gain \cite[NDCG,][]{Jarvelin2002CumulatedGE}, a well-established IR metric that places more importance on items with higher ranks.
NDCG is designed to average over multiple user annotations and queries (here corresponding to topics).

Last, we also report the raw agreement over binarized relevance.
For the human scores, we consider any documents where the fit to the category is 4 or 5 to be relevant. 
For the models, a document is considered to be relevant to a topic $k$ if its most probable topic is $k$.
The agreement is then the proportion of relevance judgments in common.

Results are in \cref{fig:topic-human-scores-all-wiki} and \cref{fig:topic-human-scores-all-bills}---of note is that \bertopic{} cluster assignments tend to have higher  agreement with human relevance judgments (binarized responses to the Fit Step), likely due to it being a clustering model.

We also report the distributions of inter-annotator correlations \emph{per topic} in \cref{fig:boxplots-by-topic}, showing that certain individual topics can have relatively high variance in human annotations.

\section{Pilot Study}\label{app:pilot}

\begin{table}
\centering
\small
\begin{tabular}{lll}
  \toprule
   & Fit Step $\alpha$ & Rank Step $\alpha$ \\
  \midrule
  \mallet{} & 0.59 (0.16) & 0.71 (0.09) \\
  \ctm{} & 0.64 (0.15) & 0.67 (0.13) \\
  \emph{Label-Derived} & 0.80 (0.13) & 0.86 (0.05) \\
  \bottomrule
  \end{tabular}
  \caption{
  Chance-corrected human--human agreement (Krippendorff's $\alpha$), averaged over the six pilot topics per model (standard deviation in parentheses) on the \texttt{Wiki} data. Each topic has between 3 and 5 annotators (the variance is due to filtering). High agreement on the synthetic labeled dataset indicates that the task is sensible.
}
\label{tab:topic-agreement-pilot}
\end{table}

\begin{figure*}[ht]
  \centering
  \includegraphics[width=\textwidth]{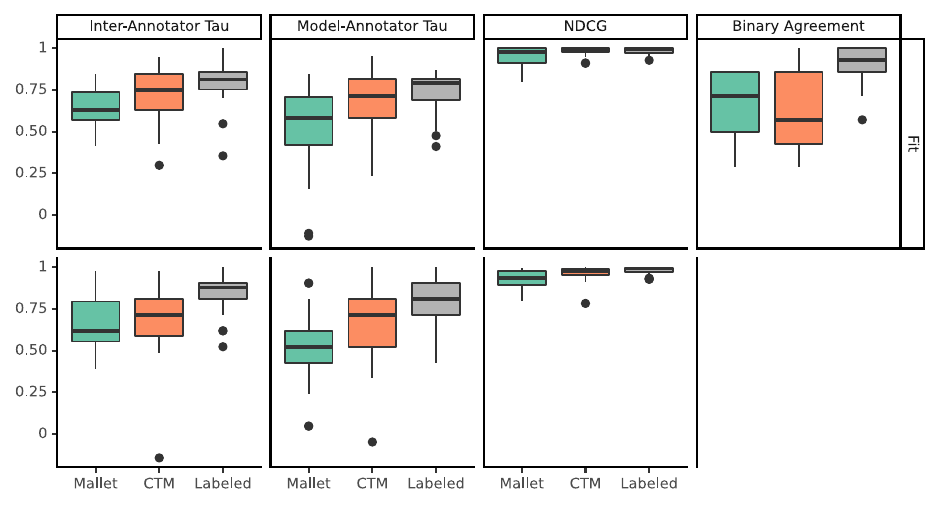}
  \caption{
    Metrics quantifying the relationship between human relevance judgments and estimated document-topic probabilities ($\bm{\theta}_k$) for two models and a synthetic upper-bound (``Labeled'', or \emph{Label-Derived} in the text), using six topics from the pilot data. From left-to-right, the metrics are inter-annotator Kendall's $\tau$,model-annotator $\tau$, relevance agreement, and NDCG. The top row of figures reports relationships with human relevance judgments (on a 1-5 scale), and the bottom row relationships with their document rankings. Boxplots report variation over topic-annotator pairs. We emphasize that the ``Labeled'' model is not a true topic model, but a synthetic supervised benchmark with access to ground-truth categories.%
  }
  \label{fig:topic-human-scores-pilot}
\end{figure*}

We first run a pilot annotation study on using the \texttt{Wiki} data on six topics from \ctm{} and \mallet{}.

To help validate the sensibility of the human evaluation protocol, we also introduce an informal upper-bound, we evaluate a synthetic ``model'' (termed \emph{Label-Derived}) using ground-truth category labels for the \texttt{Wiki} data.
For each label in data, take the documents assigned to the label $k$ and embed them (using the same embedding model as \ctm{}).
To construct a pseudo-ranking over documents for the topic, $\tilde{\bm{\theta}_k}$, we calculate the cosine similarity between the document embeddings (for all documents) and the centroid of all $k$-labeled documents.
We further correct the similarities for the $k$th label by adding 1 to all the $k$-labeled documents, ensuring that they are ranked above those that are not labeled for the document.
Synthetic top words for the topic are found by concatenating all $k$-labeled documents and computing the tf-idf for this pooled ``document''.
The result is that all \emph{exemplar documents} are known to relate to a single ground-truth label (e.g., video games).

Results show that both inter-annotator and model-annotator agreement metrics are substantially higher for the synthetic model, \cref{tab:topic-agreement-pilot}.
Of particular note are the binary agreement scores (\cref{fig:topic-human-scores-pilot}, implying that human annotators agree with a ground-truth assignment at very high rates.

The resulting annotation data is used to help tune the LLM prompts in App. \ref{app:prompting}.

\section{Notes on Agreement Metrics}\label{app:agreement}
The most straightforward way to assess relative model performance using the human annotations is to compute the chance-corrected inter-annotator agreement---indeed, this corresponds most closely to the way a manual qualitative content analysis is assessed.
A topic with high agreement across annotators is likely to be better than one with low agreement.
However, the idea is complicated by annotators only viewing one topic each.
Measures like Krippendorff's $\alpha$ \cite{krippendorff-2019-content} use the empirical distributions to estimate expected agreement when correcting for chance, so a topic with relatively high raw agreement (i.e., a very skewed distribution) may have a low value relative to what is qualitatively considered a ``good'' topic.\footnote{There is extensive literature on this issue \cite[][]{di-eugenio-glass-2004-squibs,gwet2012handbook,Xu2014InterraterAS}. Nonetheless, in the political science community, Krippendorff's $\alpha$ and Cohen's $\kappa$ remain essentially universal. As far as we can tell, this is also true more broadly in the social sciences.}%
While it is possible to average these values over topics, their occasionally counter-intuitive nature makes them less desirable for model comparison.
For a more in-depth overview of inter-annotator agreement in linguistic annotation, we refer the reader to \citet{Artstein2017}.

\section{\bertopic{} training details}\label{app:bertopic_training_details}
Although the \bertopic{} author advises against data preprocessing\footnote{\url{https://maartengr.github.io/BERTopic/faq.html\#should-i-preprocess-the-data}}, we apply the same minimal preprocessing used for training \mallet{} and \ctm{} models (tokenization and entity identification) to ensure comparable conditions (we also find that, qualitatively, topics are better after preprocessing). Contextualized embeddings are generated separately using the raw (i.e., unprocessed) text and \bertopic{}'s default embedding model (\texttt{all-MiniLM-L6-v2}). The preprocessed data and pre-calculated embeddings are then passed to the model. We use the probabilities derived from \bertopic{}'s \texttt{approximate\_distribution} function as document-topic distribution to obtain the evaluation documents.

\section{Alternative Annotator Combination Strategy for the Alt-Test}\label{app:alt_test_model_combs}
Given that each crowdworker only annotates one topic (hence seven documents), the standard application of the alt-test would be insufficiently powered.
As discussed in \cref{subsec:metrics}, must therefore combine annotators to create ``pseudo-annotators'' who appear to have annotated multiple topics, per \citet{calderon2025alternativeannotatortestllmasajudge}.

However, there are multiple ways of combining annotators.
For the results in \cref{tab:alt_test}, we combine annotators over all 24 topics per dataset; given that each topic has at least three annotators (after filtering), this produces three pseudo-annotators who each observe a set of 168 unique documents
Recall that the statistical tests are run over the computed human-human (or human-LLM) similarities across the annotated instances (i.e. documents), so higher numbers imply more power.

As an alternative that generates more pseudo-annotators (but fewer documents per annotator), we randomly combine topics per \emph{model}, rather than per dataset. This introduces more noise, as the distribution of topics viewed by each ``annotator'' is variable (e.g., some could observe all low-quality topics with high disagreement), and makes the statistical tests harder to ``pass.''
Results are in \cref{tab:alt_test_model_combs}: the $\rho$ are roughly the same as before, but statistical win rates above $0.5$ (as indicated by the $*$ and $\dagger$) are less frequent, presumably due to the higher variance and lower power.

\begin{table}
\centering
\small
\resizebox{\columnwidth}{!}{%
\input{tables/stat_test_table_model_combs}
}
\caption{
  Advantage probabilities $\rho$ from the alternative annotator test using an alternative combination method (details in App. \ref{app:alt_test_model_combs}). $\rho$ is the probability that \proxann{} is ``as good as or better than a randomly chosen human annotator'' \cite{calderon2025alternativeannotatortestllmasajudge}. Document-level scores consider annotations by document; Topic-level over all documents evaluated in the topic. $*$ indicates that win rates over humans are above 0.5, as determined by a one-sided t-test (over 10 resamples of combined annotators). $\dagger$ is the equivalent for Wilcoxon signed-rank.
}\label{tab:alt_test_model_combs}
\end{table}

\section{Additional Bills Results}

\Cref{fig:topic-human-scores-all-bills} depict evaluations on the \texttt{Bills} data, corresponding to \cref{fig:topic-human-scores-all-wiki} in the main text.

\section{Prompting details}\label{app:prompting}
\subsection{Evaluation protocol configuration}
Here, we outline our prompt engineering process used to configure the LLM-based proxy for the evaluation protocol. 
\paragraph{Label Step} We use a concise system prompt (\ref{prompt:system_prompt_q1}) to summarize the tasks and instruct the LLM to simulate human-like behavior. This is paired with an instruction prompt (\ref{prompt:prompt_q1}) that provides task-specific details, augmented with few-shot exemplars.
\paragraph{Fit and Rank Steps} Following the findings of \citet{wang2025improvingllmasajudgeinferencejudgment}, we prompt the LLM using a single instruction prompt per task, without Chain-of-Thought reasoning or few-shot exemplars (\ref{prompt:prompt_q2} and \ref{prompt:prompt_q3}). 
For the {\bf Fit Step}, we adopt a pointwise scoring approach to compute the Fit Score. Rather than relying on the LLM's most-probable token alone, we extract the log-probabilities of the top-20 tokens and interpret them as soft judgments over the Likert scale 1--5. We then compute a weighted average across the Likert candidates, using the LLM-assigned probabilities as weights.

The {\bf Rank Step} involves pairwise ranking, where the LLM is presented with pairs of texts and asked to choose which one better fits a given category. 
To ensure a fair comparison in the prompt, evaluation documents are referred to as A and B to avoid biasing the model (e.g., implying significance based on numerical identifiers). However, this approach may still introduce a preference for one letter over the other. To mitigate this, we implemented a ``both-ways'' approach, running the prompt twice for each document pair: once with the first document as A and the second as B, and vice versa (following \citealt{wang-etal-2024-large-language-models-fair,wang2025improvingllmasajudgeinferencejudgment}). As in \citet{wang2025improvingllmasajudgeinferencejudgment}, we take the probability-weighted average over the tokens in both directions before taking the final rank.

\subsection{Bradley-Terry}
After applying the {\it Rank Step} prompt to each topic on all $7 \choose 2$ combinations of evaluation document pairs, we infer the real-valued ``relatedness'' for the topic by aggregating pairwise comparisons using the Iterative Luce Spectral Ranking (ILSR) algorithm. To compute the rankings, we use the implementation from the \texttt{choix}\footnote{\url{https://choix.lum.li/en/latest/}}library,  applying the \texttt{ilsr\_pairwise} method, setting the regularization term \(\alpha\) to \(0.001\) to ensure numerical stability.

\subsection{Deployment Infrastructure}
LLama-8B models was run on an NVIDIA 4090 (24 GB RAM); all other models were run on an NVIDIA A100 (80GB RAM). 70B Models were quantized with AWQ \cite{lin2024awqactivationawareweightquantization}. vLLM was used for inference \cite{kwon2023efficientmemorymanagementlarge};\footnote{\url{https://docs.vllm.ai/}} prompting across all 24 topics takes under an hour. 

\section{Agreement evaluation}\label{appen:disagreement_evaluation}
Tables~\ref{tab:disagreement_human_tm_evaluation},~\ref{tab:example_human_llm_disagreement1} and ~\ref{tab:example_human_llm_disagreement2} contain qualitative examples to illustrate the analysis carried out in Section~\ref{sec:agreement_eval}.

\begin{table*}
  \centering
  \resizebox{\textwidth}{!}{
  \input{tables/disagreement_human_tm_evaluation}
  }
\caption{Examples of topics with the lowest human-to-human agreement (H:H), as well as topics with low and high human-to-topic model agreement (H:TM) conditioned on high H:H (defined as Krippendorff's $\alpha > 0.7$). Human-to-topic model agreement is measured using Kendall's $\tau$ on the fit scores. For each topic, we show: (1) the top eight words from the topic model, (2) the annotators' categories associated with the topic, (3) the top two exemplar documents, and (4) two evaluation documents selected from the topics with the lowest or highest H:H or H:TM agreement, depending on the condition. Both exemplar and evaluation documents display the model's $\theta_d$, and the evaluation documents additionally include the human-assigned fit rating (mean and standard deviation). Topics with low H:H tend to be broad or multi-themed, leading to disagreement in both category framing and document fit. Disagreement in low H:TM cases appears to stem from model limitations---e.g., \bertopic{}'s hard clustering approximation or \mallet{} defaulting to the best available topic despite poor fit.}
\label{tab:disagreement_human_tm_evaluation}
\end{table*}

\begin{table*}
  \centering
  \resizebox{\textwidth}{!}{
  \input{tables/example_human_llm_disagreement1}
  }
\caption{Case study of a topic with high human-to-human agreement (Krippendorff's $\alpha = 0.76$) but moderate-to-low human-to-LLM (\gptFourO) agreement (Kendall's $\tau = 0.48$), based on \bertopic{} applied to \texttt{Bills}. We show the top eight topic words, human- and LLM-generated categories, two exemplar documents, and all evaluation documents. Evaluation documents include human fit ratings (mean and standard deviation), model-assigned topic probabilities ($\theta_d$), and LLM-assigned fit scores. While the $\tau$ value may suggest poor alignment, qualitative inspection shows strong agreement between humans and the LLM on individual document fit, indicating that even moderate $\tau$ values can correspond to good topics.}
\label{tab:example_human_llm_disagreement1}
\end{table*}

\begin{table*}
  \centering
  \resizebox{\textwidth}{!}{
  \input{tables/example_human_llm_disagreement2}
  }
\caption{Same as Table~\ref{tab:example_human_llm_disagreement1}, but for a topic from \ctm{} trained on \texttt{Wiki}. This topic has very high human-to-human agreement ($\alpha = 0.98$) and moderate human-to-LLM agreement ($\tau = 0.58$). While both humans and the LLM associate the topic with baseball players, the LLM appears to emphasize notable achievements, leading to slightly lower fit scores on documents like {\it Doc ID 2932}, which reads as a biography of a less prominent player. The disagreement here is minimal but highlights how subtle differences in topic scope can influence agreement.}
\label{tab:example_human_llm_disagreement2}
\end{table*}

\section{Additional Topic Ranking Results}\label{app:model_ranks}
\begin{figure*}
    \centering
    \includegraphics[width=\textwidth]{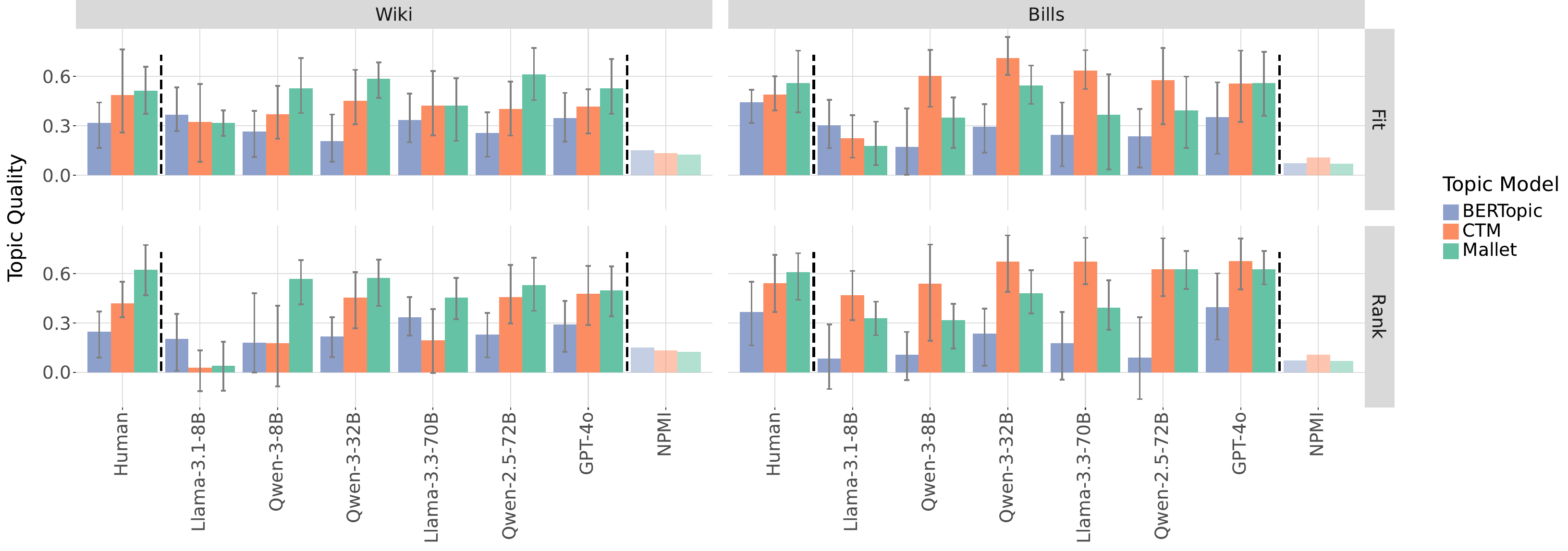}
    \caption{Model rankings based on $\fittauHTM{}$/ $\ranktauHTM{}$ (correlation of human scores with document-topic probabilities), $\fittauLMTM$/$\ranktauLMTM{}$ (correlation of LLM scores with document-topic probabilities), and NPMI coherence. Correlations are computed using Kendall's \(\tau\). Error bars for Human and LLM metrics are 95\% bootstrapped confidence intervals, resampled over topics. \ctm{} performs best on \texttt{Bills}, while \mallet{} leads on \texttt{Wiki}. LLM-based metrics align well with human judgments, unlike NPMI. Rankings are consistent across LLMs.}
    \label{fig:all_metrics_models_comparison_tau}
\end{figure*}

\begin{table}[!t]
\centering
\label{tab:topline_table_binarized}
\resizebox{\columnwidth}{!}{%

\input{tables/topline_table_tau_binarized}
}
\caption{Relationship between automated and human topic rankings given \emph{binary} topic assignments. Cells show Kendall's $\tau$ between metrics: $\textsc{Fit/}\ranktauHTM{}$ correlates human scores to binary assignments: $\mathbb{I}\left[\text{argmax}\left(\bm{\theta}_k\right) = k\right]$. $\textsc{Fit/}\ranktauLMTM{}$ correlates \proxann{} to binary assignments. Values are bootstrapped means and standard deviations (resampling over topics). The \emph{Human} row reflects leave-one-out inter-annotator correlations, serving as a reference.}
\end{table}

\cref{fig:all_metrics_models_comparison_tau} reports topic model rankings using our automated metrics, $\fittauLMTM$, $\ranktauLMTM{}$.

We also report how well topic rankings correlate when using binary assignments from the topic model---that is, whether the topic is the most-probable for that document, rather than the original real-valued $\bm{\theta}_k$. Generally, the correlations are higher (\cref{tab:topline_table_binarized}), suggesting the use of assignments may be preferable for topic rankings.

\section{User Interface}\label{app:interface}
\begin{figure*}[ht]
    \centering
    \includegraphics[width=0.75\textwidth]{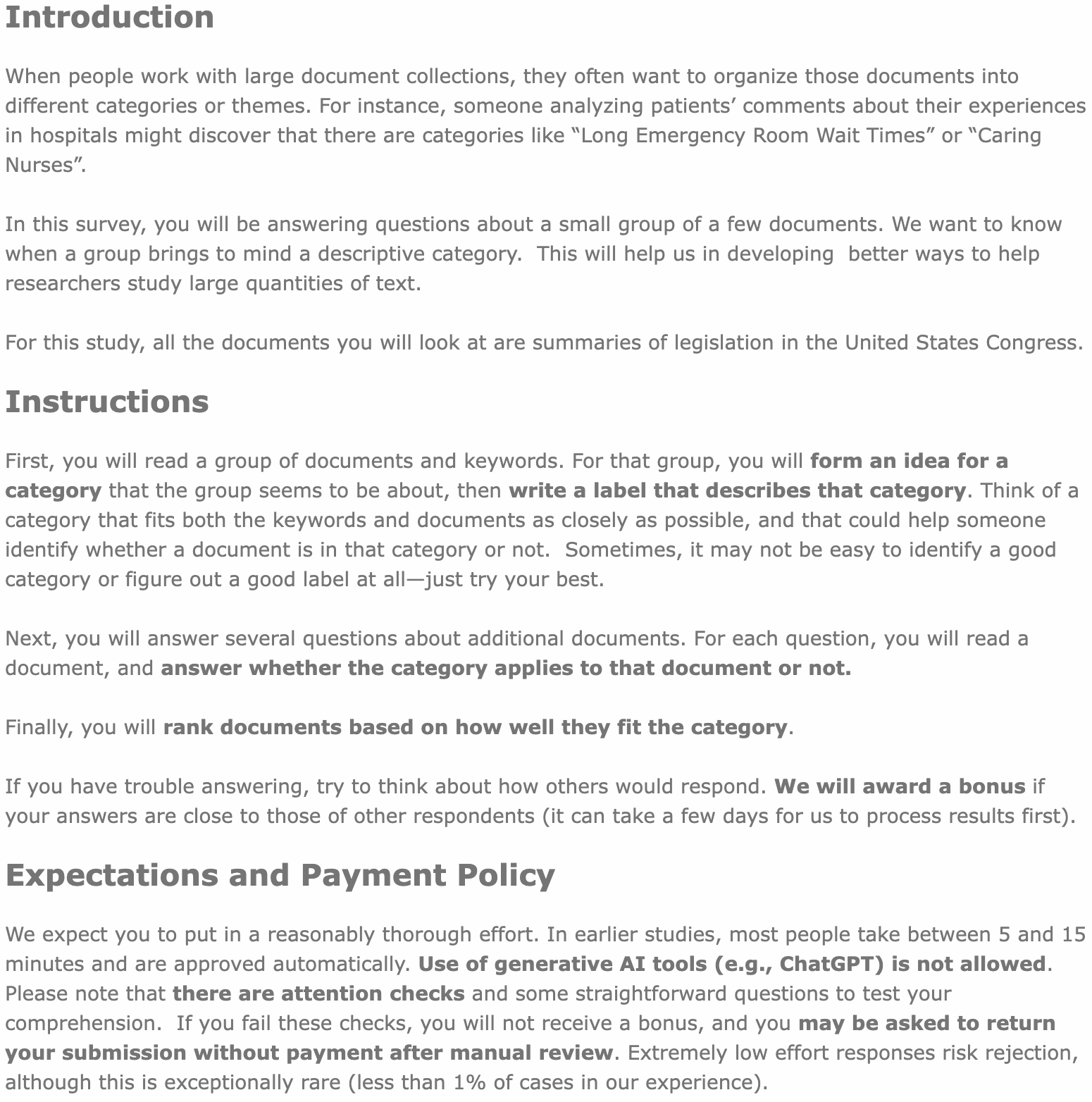}
    \caption{Instructions for the human annotation protocol.}
\label{fig:topic-interface-instructions}
\end{figure*}
\begin{figure*}[ht]
    \centering
    \includegraphics[width=0.75\textwidth]{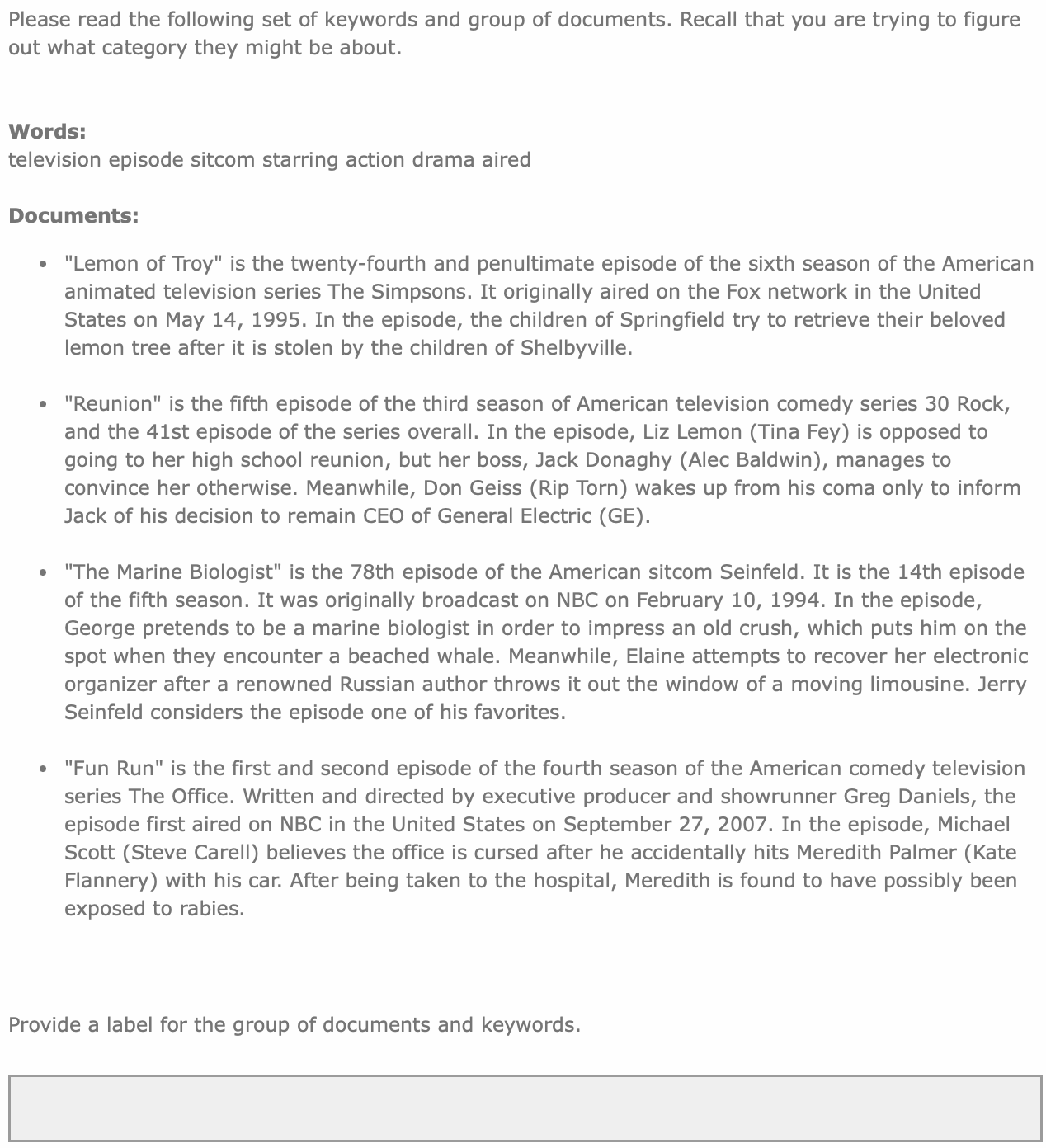}
    \caption{Label Step. Category identification in the human annotation protocol for the practice question.}
\label{fig:topic-interface-label}
\end{figure*}
\begin{figure*}[ht]
    \centering
    \includegraphics[width=0.75\textwidth]{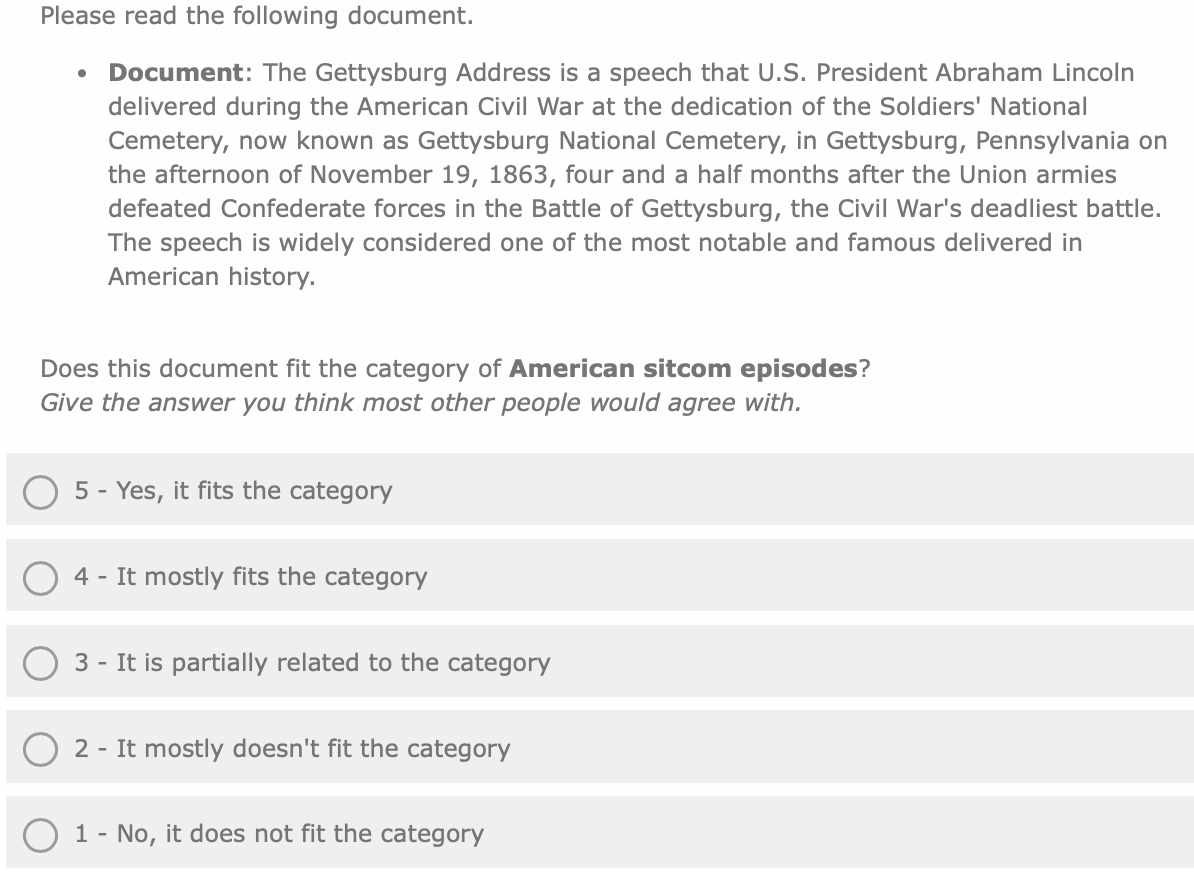}
    \caption{Fit Step. Relevance judgment in the human annotation protocol for the practice question.}
\label{fig:topic-interface-fit}
\end{figure*}
\begin{figure*}[ht]
    \centering
    \includegraphics[width=0.75\textwidth]{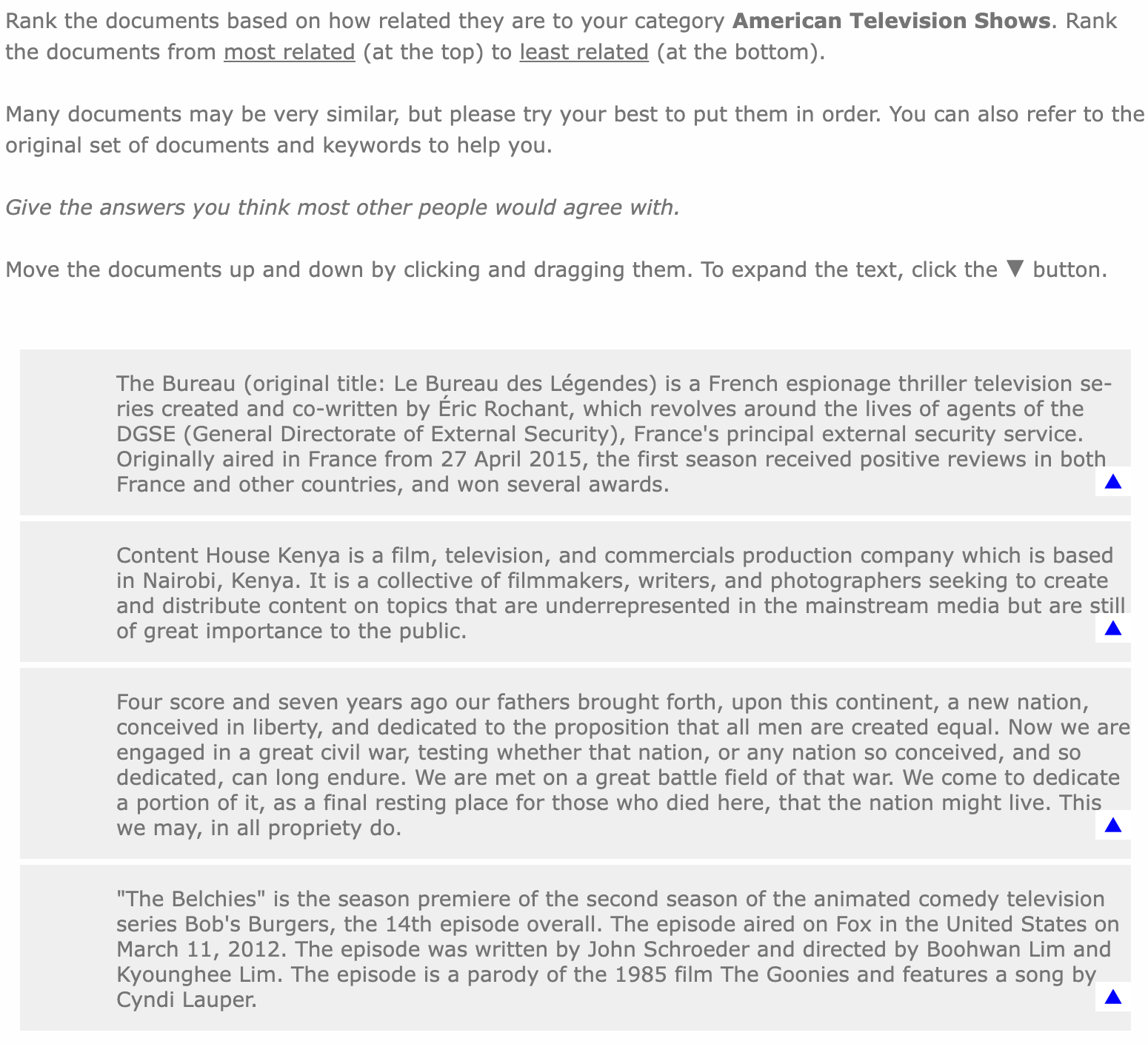}
    \caption{Rank Step. Representativeness ranking in the human annotation protocol for the practice question.}
\label{fig:topic-interface-rank}
\end{figure*}
\begin{figure*}[ht]
    \centering
    \includegraphics[width=0.75\textwidth]{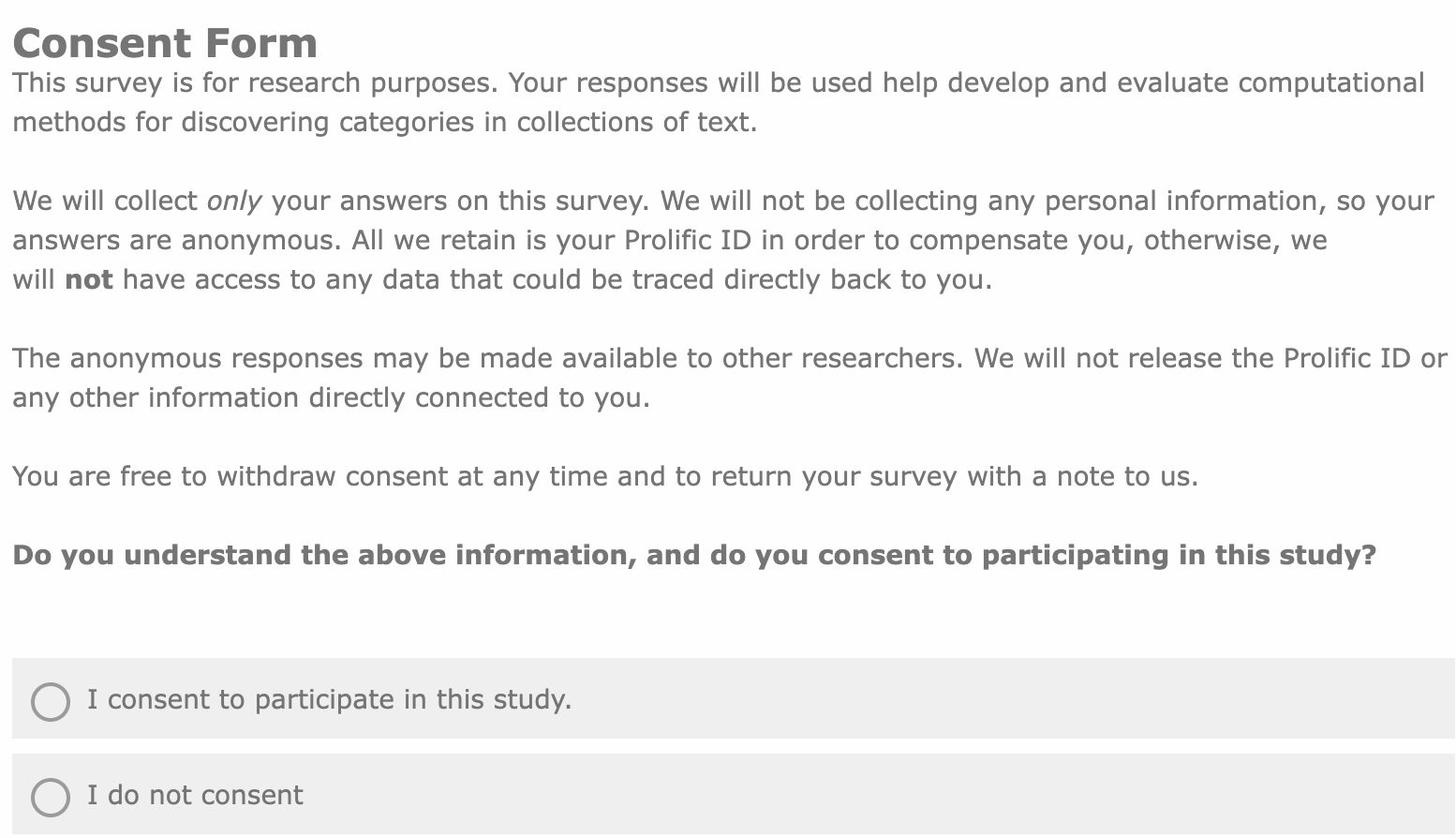}
    \caption{Consent page (shown at beginning)}
\label{fig:topic-interface-consent}
\end{figure*}

\Cref{fig:topic-interface-instructions,fig:topic-interface-label,fig:topic-interface-fit,fig:topic-interface-rank} are screenshots of the annotation interface presented to users. \Cref{fig:topic-interface-consent} is the consent page shown at the start.

\onecolumn
\section{Prompt templates}
\begin{prompt}[title={\thetcbcounter: Category Identification (Label Step, System Prompt)}, label=prompt:system_prompt_q1]
\begin{lstlisting}
You are a helpful AI assistant tasked with creating descriptive labels for a set of keywords and a group of documents, each focused on a common topic, as similar as possible to how a human would do. The goal is to provide meaningful, concise labels that capture the central theme or key concepts represented by the keywords and documents.
\end{lstlisting}
\end{prompt}

\begin{prompt}[title={\thetcbcounter: Category Identification (Label Step, Instruction Prompt)}, label=prompt:prompt_q1]
\begin{lstlisting}
You will be provided with a set of keywords and a group of documents, each centered around a common topic. Your task is to analyze both the keywords and the content of the documents to create a clear, concise label that accurately reflects the overall theme they share.

Task Breakdown:
1. Examine the Keywords: Use the keywords as clues to identify the general subject area or themes present in the documents.
2. Review the Documents: Skim the summaries provided to understand their main ideas and any recurring elements.
3. Generate a Label: Based on the keywords and document content, come up with a single label that best describes the topic connecting all the documents.

Examples:
--------
{}

#########

KEYWORDS: {}
DOCUMENTS: {}
Based on the keywords and document content, come up with a single category that best describes the topic connecting all the documents. Return just the category.
CATEGORY:
\end{lstlisting}
\end{prompt}

\begin{prompt}[title={\thetcbcounter: Relevance Judgment (Fit Step)}, label=prompt:prompt_q2]
\begin{lstlisting}
Please act as an impartial judge and assign an integer score from 1 to 5 indicating how well the DOCUMENT fits the given CATEGORY. Do not provide any reasoning or explanation

[[ ## CATEGORY ## ]]
{category}

[[ ## DOCUMENT ## ]]
{document}
\end{lstlisting}
\end{prompt}

\begin{prompt}[title={\thetcbcounter: Representativeness Pairwise Ranking (Rank Step)}, label=prompt:prompt_q3]
\begin{lstlisting}
Please act as an impartial judge and determine which of the two documents (A or B) is more closely related to the given CATEGORY. Avoid any positional bias, and ensure that the order in which the documents are presented does not influence your decision. Output your verdict strictly using this format: 'A' if DOCUMENT_A is more closely related to the CATEGORY, or 'B' if DOCUMENT_B is more closely related.

[[ ## CATEGORY ## ]]
{category}

[[ ## DOCUMENT_A ## ]]
{doc_a}

[[ ## DOCUMENT_B ## ]]
{doc_b}
\end{lstlisting}
\end{prompt}

\twocolumn

\end{document}

%% file: tables/topics_examples_table.tex
\begin{tabular}{lll}
  \toprule
  \mallet{} & CTM & BERTopic \\
  \midrule
  \emph{season game games home runs} & \emph{career hit games season league} & \emph{yard season team yards league} \\
  \textcolor{customblue}{Major League Baseball Players and History} &  \textcolor{customblue}{Major League Baseball Players and Achievements} &  \textcolor{customblue}{Sports and Athletics} \\
  Professional baseball players & Former MLB players  & Professional Basketball and Baseball Players \\
  American professional baseball players & American baseball league & American sports and their associated famous sportsmen \\
  Baseball knowledge hub & Professional baseball facts and figures  & Sports champions \\
  \cline{1-3}
  \emph{act consumer credit employee card} & \emph{fuel credit revenue internal property} & \emph{vehicle recorder motor retrieved retrieval} \\
  \textcolor{customblue}{Labor and Employment Legislation} &  \textcolor{customblue}{Renewable Energy Tax Credits and Incentives} &  \textcolor{customblue}{Vehicle Data Privacy and Ownership Rights} \\
  Individual Protection Laws    & Renewable energy tax and biofuel             & Vehicle owner protections \\
  Labor Laws and Protections    & Energy tax credits, Alternative fuel credits & Automobile Ownership Legislation \\
  Proposed employee protections & Energy Tax Policy                            & Vehicle Owner and Safety Legislation \\
  \bottomrule
\end{tabular}

%% file: tables/stat_test_table.tex
\begin{tabular}{lllll}
\toprule
  & \multicolumn{2}{c}{Document-Level $\rho$} & \multicolumn{2}{c}{Topic-Level $\rho$} \\
\cmidrule(lr){2-3} \cmidrule(lr){4-5}
   & Fit & Rank & Fit & Rank \\
\midrule
& \multicolumn{4}{c}{\texttt{Wiki}}\\
\cmidrule(lr){2-5}
\gptFourO{} & $0.56^{{*}\dagger}$ & $0.68^{{*}\dagger}$ & $0.66^{\dagger}$ & $0.55^{\dagger}$ \\
\llamaThreeOneEightB{} & $0.22$ & $0.36$ & $0.05$ & $0.11$ \\
\llamaThreeThreeSeventyB{} & $0.57^{{*}\dagger}$ & $0.67^{{*}\dagger}$ & $0.58^{\dagger}$ & $0.50^{\dagger}$ \\
\qwenThreeEightB{} & $0.56^{{*}\dagger}$ & $0.58^{\dagger}$ & $0.46$ & $0.39$ \\
\qwenThreeThirtyTwoB{} & $0.55^{{*}\dagger}$ & $0.63^{\dagger}$ & $0.47$ & $0.42$ \\
\qwenTwoFiveSeventyTwoB{} & $0.52^{\dagger}$ & $0.68^{{*}\dagger}$ & $0.66^{\dagger}$ & $0.46$ \\
\midrule
& \multicolumn{4}{c}{\texttt{Bills}}\\
\cmidrule(lr){2-5}
\gptFourO{} & $0.65^{{*}\dagger}$ & $0.71^{{*}\dagger}$ & $0.77^{{*}\dagger}$ & $0.75^{{*}\dagger}$ \\
\llamaThreeOneEightB{} & $0.30$ & $0.53^{\dagger}$ & $0.14$ & $0.44$ \\
\llamaThreeThreeSeventyB{} & $0.66^{{*}\dagger}$ & $0.67^{{*}\dagger}$ & $0.70^{{*}\dagger}$ & $0.60^{\dagger}$ \\
\qwenThreeEightB{} & $0.66^{{*}\dagger}$ & $0.57^{\dagger}$ & $0.80^{{*}\dagger}$ & $0.43$ \\
\qwenThreeThirtyTwoB{} & $0.67^{{*}\dagger}$ & $0.68^{{*}\dagger}$ & $0.74^{{*}\dagger}$ & $0.70^{{*}\dagger}$ \\
\qwenTwoFiveSeventyTwoB{} & $0.61^{{*}\dagger}$ & $0.71^{{*}\dagger}$ & $0.78^{{*}\dagger}$ & $0.65^{\dagger}$ \\
\bottomrule
\end{tabular}

%% file: tables/topline_table.tex
\begin{tabular}{@{}p{2.4cm}cccc@{}}
\toprule
\multirow{2}{*}{} & \multicolumn{2}{c}{$\tau\left(\fittauHTM,\cdot \right)$} & \multicolumn{2}{c}{$\tau\left(\ranktauHTM,\cdot \right)$} \\
\cmidrule(lr){2-3}\cmidrule(lr){4-5}
& \texttt{Wiki} & \texttt{Bills} & \texttt{Wiki} & \texttt{Bills} \\
\midrule
NPMI & -0.15 (0.14) & 0.01 (0.10) & -0.18 (0.10) & -0.02 (0.12) \\
\midrule
\multicolumn{2}{l}{$\textsc{fit/}\ranktauLMTM$} & & & \\
\cmidrule(lr){1-1}
\llamaThreeOneEightB{}   & 0.19 (0.18) & 0.16 (0.18) & -0.35 (0.14)  & 0.15 (0.14) \\
\qwenThreeEightB{}     & 0.35 (0.16) & 0.12 (0.16) & 0.33 (0.16)  & 0.28 (0.13) \\
\qwenThreeThirtyTwoB{}     & 0.20 (0.18) & \textbf{0.34 (0.11)} & \textbf{0.51 (0.11)}  & \textbf{0.30 (0.13)} \\
\llamaThreeThreeSeventyB{}  & 0.41 (0.14) & 0.26 (0.15) & 0.36 (0.13)  & 0.19 (0.13) \\
\qwenTwoFiveSeventyTwoB{}   & \textbf{0.48 (0.13)} & 0.22 (0.17) & 0.36 (0.12) & 0.21 (0.15) \\
\gptFourO{}         & 0.22 (0.13) & 0.31 (0.13) & 0.27 (0.14)  & 0.29 (0.11) \\
\midrule
\multicolumn{2}{l}{$\textsc{fit/}\ranktauHTM$} & & & \\
\cmidrule(lr){1-1}
\emph{Human} & 0.41 (0.09) & 0.09 (0.14) & 0.34 (0.09)  & 0.18 (0.12) \\
\bottomrule
\end{tabular}

%% file: tables/stat_test_table_model_combs.tex
\begin{tabular}{lllll}
\toprule
  & \multicolumn{2}{c}{Document-Level $\rho$} & \multicolumn{2}{c}{Topic-Level $\rho$} \\
\cmidrule(lr){2-3} \cmidrule(lr){4-5}
   & Fit & Rank & Fit & Rank \\
\midrule
& \multicolumn{4}{c}{\texttt{wiki}}\\
\cmidrule(lr){2-5}
\gptFourO{} & $0.58^{\dagger}$ & $0.68^{\dagger}$ & $0.67$ & $0.55$ \\
\llamaThreeOneEightB{} & $0.22$ & $0.37$ & $0.04$ & $0.12$ \\
\llamaThreeThreeSeventyB{} & $0.58^{\dagger}$ & $0.67^{\dagger}$ & $0.58$ & $0.48$ \\
\qwenThreeEightB{} & $0.58^{\dagger}$ & $0.58$ & $0.46$ & $0.38$ \\
\qwenThreeThirtyTwoB{} & $0.57^{\dagger}$ & $0.63$ & $0.51$ & $0.42$ \\
\qwenTwoFiveSeventyTwoB{} & $0.53^{\dagger}$ & $0.68^{\dagger}$ & $0.67$ & $0.46$ \\
\midrule
& \multicolumn{4}{c}{\texttt{bills}}\\
\cmidrule(lr){2-5}
\gptFourO{} & $0.65^{{*}\dagger}$ & $0.71^{{*}\dagger}$ & $0.77$ & $0.76$ \\
\llamaThreeOneEightB{} & $0.30$ & $0.53^{\dagger}$ & $0.14$ & $0.44$ \\
\llamaThreeThreeSeventyB{} & $0.66^{{*}\dagger}$ & $0.67^{\dagger}$ & $0.69$ & $0.60$ \\
\qwenThreeEightB{} & $0.66^{{*}\dagger}$ & $0.57^{\dagger}$ & $0.80^{{*}}$ & $0.43$ \\
\qwenThreeThirtyTwoB{} & $0.67^{{*}\dagger}$ & $0.68^{\dagger}$ & $0.75$ & $0.71$ \\
\qwenTwoFiveSeventyTwoB{} & $0.61^{{*}\dagger}$ & $0.70^{\dagger}$ & $0.79$ & $0.65$ \\
\bottomrule
\end{tabular}

%% file: tables/disagreement_human_tm_evaluation.tex
\renewcommand{\arraystretch}{1.4}
\begin{tabular}{ >{\centering\arraybackslash}p{3cm} p{5cm} p{4.5cm} p{4.5cm} p{4.5cm} p{4.5cm} }
\toprule
\textbf{Topic Words} & \textbf{Categories} & \textbf{Exemplar Document 1} & \textbf{Exemplar document 2} & \textbf{Evaluation Document 1} & \textbf{Evaluation Document 2} \\
\midrule
\rowcolor{gray!15} \multicolumn{6}{c}{\textbf{Lowest H:H ($\alpha < 0.7$)}} \\
\multicolumn{6}{c}{\textit{BERTopic on \texttt{\texttt{Bills}} ($\alpha = -0.16$)}} \\
student, school, students, education, schools, leas, higher, ihes & \begin{minipage}[t]{\linewidth}
\begin{itemize}
\setlength{\itemsep}{0pt}
\setlength{\parskip}{0pt}
\item Aiding children in school to receive a proper, well informed, education in school and post secondary.
\item Educational reform and students' welfare
\item K-12 federal education legislation
\item High School Student Initiative
\end{itemize}
\end{minipage} & \begin{minipage}[t]{\linewidth}
\textit{\textbf{Doc ID} 1947: $\theta_d=1.00$} \\
\ul{Text:} English Language Instruction Improvement Act of 2007 - Amends title III (Language Instruction for Limited English Proficient and Immigrant Students) [...]
\end{minipage} & \begin{minipage}[t]{\linewidth}
\textit{\textbf{Doc ID} 1375: $\theta_d=1.00$} \\
\ul{Text:} Scholarships for Opportunity and Results Reauthorization Act or the SOAR Reauthorization Act

This bill amends the District of Columbia Code to [...]
\end{minipage} & \begin{minipage}[t]{\linewidth}
\textit{\textbf{Doc ID} 197: $\text{Human Fit}=2.50\pm1.66$, $\theta_d=1.00$} \\
\ul{Text:} Community College Partnership Act of 2007 - Amends the Higher Education Act of 1965 to establish a Community College Opportunity program to help [...]
\end{minipage} & \begin{minipage}[t]{\linewidth}
\textit{\textbf{Doc ID} 1387: $\text{Human Fit}=2.50\pm1.66$, $\theta_d=0.33$} \\
\ul{Text:} Dynamic Repayment Act of 2016 

This bill amends the Higher Education Act of 1965 to replace several existing federal student loan programs with a [...]
\end{minipage} \\
\addlinespace
\multicolumn{6}{c}{\textit{\mallet{} on \texttt{Bills}($\alpha = -0.15$)}} \\
act, funds, year, federal, fiscal, fund, amounts, state & \begin{minipage}[t]{\linewidth}
\begin{itemize}
\setlength{\itemsep}{0pt}
\setlength{\parskip}{0pt}
\item Guidelines for approving Congressional Budget Expenditures
\item Specific government spending limitation \texttt{Bills}
\item Legislative Acts
\item Federal American Fiscal Funding Acts 
\end{itemize}
\end{minipage} & \begin{minipage}[t]{\linewidth}
\textit{\textbf{Doc ID} 2946: $\theta_d=0.99$} \\
\ul{Text:} Realize America's Maritime Promise Act or the RAMP Act - Requires the total budget resources for expenditures from the Harbor Maintenance Trust Fund [...]
\end{minipage} & \begin{minipage}[t]{\linewidth}
\textit{\textbf{Doc ID} 9315: $\theta_d=0.82$} \\
\ul{Text:} Midshipmen Education Certainty Act 

 Makes appropriations available each fiscal year for operations of the U.S.
\end{minipage} & \begin{minipage}[t]{\linewidth}
\textit{\textbf{Doc ID} 11330: $\text{Human Fit}=3.50\pm1.12$, $\theta_d=0.00$} \\
\ul{Text:} (This measure has not been amended since the Conference Report was filed in the House on June 29, 2010.
\end{minipage} & \begin{minipage}[t]{\linewidth}
\textit{\textbf{Doc ID} 14524: $\text{Human Fit}=3.50\pm1.12$, $\theta_d=0.17$} \\
\ul{Text:} Target Practice and Marksmanship Training Support Act 

 This bill amends the Pittman-Robertson Wildlife Restoration Act to facilitate the [...]
\end{minipage} \\
\addlinespace
\midrule
\rowcolor{gray!15} \multicolumn{6}{c}{\textbf{Low H:TM (High H:H $\alpha > 0.7$)}} \\
\multicolumn{6}{c}{\textit{BERTopic on \texttt{Wiki}($\alpha = 0.72$)}} \\
breed, horses, horse, breeds, arabian, dogs, dog, bred & \begin{minipage}[t]{\linewidth}
\begin{itemize}
\setlength{\itemsep}{0pt}
\setlength{\parskip}{0pt}
\item Animal Breeds
\item animal breeds, horse and dog breeds
\item Animal breeds
\item Domestic animals
\item Breeds horses dogs historical working
\item Native four legged animals
\end{itemize}
\end{minipage} & \begin{minipage}[t]{\linewidth}
\textit{\textbf{Doc ID} 5378: $\theta_d=1.00$} \\
\ul{Text:} Carolina Marsh Tacky = The Carolina Marsh Tacky or Marsh Tacky is a rare breed of horse , native to South Carolina .
\end{minipage} & \begin{minipage}[t]{\linewidth}
\textit{\textbf{Doc ID} 5362: $\theta_d=1.00$} \\
\ul{Text:} Clumber Spaniel = The Clumber Spaniel is a breed of dog of the spaniel type , developed in the United Kingdom .
\end{minipage} & \begin{minipage}[t]{\linewidth}
\textit{\textbf{Doc ID} 9874: $\text{Human Fit}=4.57\pm0.73$, $\theta_d=0.17$} \\
\ul{Text:} Paynter ( horse ) = Paynter ( foaled March 4 , 2009 ) is an American-bred Thoroughbred racehorse notable for a promising three-year-old racing [...]
\end{minipage} & \begin{minipage}[t]{\linewidth}
\textit{\textbf{Doc ID} 5298: $\text{Human Fit}=4.57\pm0.73$, $\theta_d=0.34$} \\
\ul{Text:} Field Spaniel = The Field Spaniel is a medium-sized breed dog of the spaniel type .
\end{minipage} \\
\addlinespace
\multicolumn{6}{c}{\textit{\texttt{Wiki} on \mallet{} ($\alpha = 0.81$)}} \\
species, shark, long, sharks, females, fish, found, birds & \begin{minipage}[t]{\linewidth}
\begin{itemize}
\setlength{\itemsep}{0pt}
\setlength{\parskip}{0pt}
\item Sharks in the pacific ocean
\item Description of marine predators
\item Shark species descriptions and habitats 
\item shark species
\item Sea wildlife species
\end{itemize}
\end{minipage} & \begin{minipage}[t]{\linewidth}
\textit{\textbf{Doc ID} 170: $\theta_d=0.99$} \\
\ul{Text:} Banded houndshark = The banded houndshark ( Triakis scyllium ) is a species of houndshark , in the family Triakidae , common in the northwestern [...]
\end{minipage} & \begin{minipage}[t]{\linewidth}
\textit{\textbf{Doc ID} 190: $\theta_d=0.98$} \\
\ul{Text:} Coral catshark = The coral catshark ( Atelomycterus marmoratus ) is a species of catshark , and part of the family Scyliorhinidae .
\end{minipage} & \begin{minipage}[t]{\linewidth}
\textit{\textbf{Doc ID} 224: $\text{Human Fit}=3.00\pm0.63$, $\theta_d=0.33$} \\
\ul{Text:} Fish = A fish is any member of a paraphyletic group of organisms that consist of all gill-bearing aquatic craniate animals that lack limbs with [...]
\end{minipage} & \begin{minipage}[t]{\linewidth}
\textit{\textbf{Doc ID} 204: $\text{Human Fit}=3.00\pm0.63$, $\theta_d=1.00$} \\
\ul{Text:} Pyjama shark = The pyjama shark or striped catshark ( Poroderma africanum ) is a species of catshark , and part of the family Scyliorhinidae , [...]
\end{minipage} \\
\addlinespace
\midrule
\rowcolor{gray!15} \multicolumn{6}{c}{\textbf{High H:TM (High H:H $\alpha > 0.7$)}} \\
\multicolumn{6}{c}{\textit{CTM on \texttt{Wiki} ($\alpha = 0.95$, $\tau = 0.85$)}} \\
daily, trunkline, roadway, national\_highway\_system, travels, designated, surveys, entire & \begin{minipage}[t]{\linewidth}
\begin{itemize}
\setlength{\itemsep}{0pt}
\setlength{\parskip}{0pt}
\item highway routes traffic traveling
\item State Highways
\item State Highway Rules
\item American highway routes
\end{itemize}
\end{minipage} & \begin{minipage}[t]{\linewidth}
\textit{\textbf{Doc ID} 716: $\theta_d=0.37$} \\
\ul{Text:} Ohio State Route 85 = State Route 85 ( SR 85 , OH 85 ) is an east – west state highway in the northeastern Ohio .
\end{minipage} & \begin{minipage}[t]{\linewidth}
\textit{\textbf{Doc ID} 5768: $\theta_d=0.37$} \\
\ul{Text:} Delaware Route 42 = Delaware Route 42 ( DE 42 ) is a state highway in Kent County , Delaware .
\end{minipage} & \begin{minipage}[t]{\linewidth}
\textit{\textbf{Doc ID} 598: $\text{Human Fit}=5.00\pm0.00$, $\theta_d=0.35$} \\
\ul{Text:} K-22 ( Kansas highway ) = K-22 is a 3.087-mile-long ( 4.968 km ) highway in the U.S.
\end{minipage} & \begin{minipage}[t]{\linewidth}
\textit{\textbf{Doc ID} 5696: $\text{Human Fit}=5.00\pm0.00$, $\theta_d=0.42$} \\
\ul{Text:} Delaware Route 44 = Delaware Route 44 ( DE 44 ) is a state highway in Kent County , Delaware .
\end{minipage} \\
\addlinespace
\multicolumn{6}{c}{\textit{CTM on \texttt{Wiki} ($\alpha = 0.98$, $\tau = 0.82$)}} \\
career, hit, games, season, league, baseball, major\_league\_baseball, signed & \begin{minipage}[t]{\linewidth}
\begin{itemize}
\setlength{\itemsep}{0pt}
\setlength{\parskip}{0pt}
\item Former MLB players
\item American baseball league
\item Professional baseball facts and figures
\end{itemize}
\end{minipage} & \begin{minipage}[t]{\linewidth}
\textit{\textbf{Doc ID} 2943: $\theta_d=0.61$} \\
\ul{Text:} Brian Wilson ( baseball ) = Brian Patrick Wilson ( born March 16 , 1982 ) is a former American professional baseball relief pitcher .
\end{minipage} & \begin{minipage}[t]{\linewidth}
\textit{\textbf{Doc ID} 2928: $\theta_d=0.59$} \\
\ul{Text:} Byron McLaughlin = Byron Scott McLaughlin ( born September 29 , 1955 ) is an American retired professional baseball player , alleged counterfeit [...]
\end{minipage} & \begin{minipage}[t]{\linewidth}
\textit{\textbf{Doc ID} 3001: $\text{Human Fit}=5.00\pm0.00$, $\theta_d=0.61$} \\
\ul{Text:} Johnny Evers = John Joseph Evers ( July 21 , 1881 – March 28 , 1947 ) was an American professional baseball second baseman and manager .
\end{minipage} & \begin{minipage}[t]{\linewidth}
\textit{\textbf{Doc ID} 2870: $\text{Human Fit}=5.00\pm0.00$, $\theta_d=0.74$} \\
\ul{Text:} Jon Lieber = Jonathan Ray Lieber ( born April 2 , 1970 ) is a former Major League Baseball ( MLB ) pitcher .
\end{minipage} \\
\addlinespace
\bottomrule
\end{tabular}

%% file: tables/example_human_llm_disagreement1.tex
\renewcommand{\arraystretch}{1.4}
\begin{tabular}{p{5cm}p{6cm}p{5cm}}
\hline
\rowcolor{gray!15}
\multicolumn{3}{c}{\textbf{BERTopic on \texttt{Bills} (\(\alpha = 0.76,\ \tau = 0.48\))}} \\
\textbf{Topic Words} & \textbf{Human Categories} & \textbf{LLM Category} \\
spirits, distilled, beer, wine, excise, brewers, cider, wines & \begin{minipage}[t]{\linewidth}
\begin{itemize}
\setlength{\itemsep}{2pt}
\setlength{\parskip}{0pt}
\item Distilled Goods Legislation
\item Alcohol Internal Revenue Code
\item LEGAL INVOICE
\item tax reform on alcohol products
\end{itemize}
\end{minipage} & Alcoholic Beverage Taxation and Regulation \\ \addlinespace \midrule
\multicolumn{3}{l}{\textbf{Exemplar Documents}:} \\
\multicolumn{3}{l}{\begin{minipage}[t]{\linewidth}
\begin{itemize}
\setlength{\itemsep}{4pt}
\setlength{\parskip}{0pt}
\item \begin{minipage}[t]{\linewidth}
\textit{\textbf{Doc ID} 7046: $\theta_d=1.00$} \\
\ul{Text:} Amends the Internal Revenue Code to exclude from determination of the production period for distilled spirits any period allocated to the natural [...]
\end{minipage}
\item \begin{minipage}[t]{\linewidth}
\textit{\textbf{Doc ID} 7886: $\theta_d=1.00$} \\
\ul{Text:} Aged Distilled Spirits Competitiveness Act - Amends the Internal Revenue Code to exclude the aging period from the production period for distilled [...]
\end{minipage}
\item \begin{minipage}[t]{\linewidth}
\textit{\textbf{Doc ID} 7487: $\theta_d=1.00$} \\
\ul{Text:} Reinvesting in U.S.
\end{minipage}
\end{itemize}
\end{minipage}} \\ \addlinespace \midrule
\multicolumn{3}{l}{\textbf{Evaluation Documents}:} \\
\multicolumn{3}{l}{\begin{minipage}[t]{\linewidth}
\begin{itemize}
\setlength{\itemsep}{4pt}
\setlength{\parskip}{0pt}
\item \begin{minipage}[t]{\linewidth}
\textit{\textbf{Doc ID} 6797: $\text{Human Fit}=5.00\pm0.00$, $\text{LLM Fit}=4.77$, $\theta_d=1.00$} \\
\ul{Text:} Amends the Internal Revenue Code to reduce from $18 to $9 (its pre-1991 level) the per-barrel tax on beer.
\end{minipage}
\item \begin{minipage}[t]{\linewidth}
\textit{\textbf{Doc ID} 7335: $\text{Human Fit}=4.75\pm0.43$, $\text{LLM Fit}=4.13$, $\theta_d=1.00$} \\
\ul{Text:} Amends the Internal Revenue Code to exclude from determination of the production period for distilled spirits any period allocated to the natural [...]
\end{minipage}
\item \begin{minipage}[t]{\linewidth}
\textit{\textbf{Doc ID} 7916: $\text{Human Fit}=4.75\pm0.43$, $\text{LLM Fit}=5.00$, $\theta_d=0.38$} \\
\ul{Text:} Cider Industry Deserves Equal Regulation Act or the CIDER Act 

Amends the Internal Revenue to revise the definition of "hard cider," for purposes [...]
\end{minipage}
\item \begin{minipage}[t]{\linewidth}
\textit{\textbf{Doc ID} 7576: $\text{Human Fit}=5.00\pm0.00$, $\text{LLM Fit}=4.98$, $\theta_d=0.31$} \\
\ul{Text:} Brewers Excise and Economic Relief Act of 2011 - Amends the Internal Revenue Code to: (1) reduce from $18 to $9 ( the pre-1991 level) the per-barrel [...]
\end{minipage}
\item \begin{minipage}[t]{\linewidth}
\textit{\textbf{Doc ID} 7073: $\text{Human Fit}=2.00\pm1.73$, $\text{LLM Fit}=1.94$, $\theta_d=0.28$} \\
\ul{Text:} Amends the Internal Revenue Code to increase the excise tax rate on small cigars to \$19.50 per thousand (the same rate as for small cigarettes).
\end{minipage}
\item \begin{minipage}[t]{\linewidth}
\textit{\textbf{Doc ID} 7927: $\text{Human Fit}=4.75\pm0.43$, $\text{LLM Fit}=4.99$, $\theta_d=0.17$} \\
\ul{Text:} Distillery Excise Tax Reform Act of 2015

Amends the Internal Revenue Code to allow a reduction (from $13.50 to $2.70 on each proof gallon produced [...]
\end{minipage}
\item \begin{minipage}[t]{\linewidth}
\textit{\textbf{Doc ID} 16: $\text{Human Fit}=1.00\pm0.00$, $\text{LLM Fit}=1.01$, $\theta_d=0.00$} \\
\ul{Text:} Medicare Part D Drug Class Protection Act of 2007 - Amends part D (Voluntary Prescription Drug Benefit Program) of title XVIII (Medicare) of the [...]
\end{minipage}
\end{itemize}
\end{minipage}} \\
\addlinespace
\bottomrule
\end{tabular}

%% file: tables/example_human_llm_disagreement2.tex
\renewcommand{\arraystretch}{1.4}
\begin{tabular}{p{5cm}p{6cm}p{5cm}}
\hline
\rowcolor{gray!15}
\multicolumn{3}{c}{\textbf{CTM on \texttt{Wiki} (\(\alpha = 0.98,\ \tau = 0.58\))}} \\
\textbf{Topic Words} & \textbf{Human Categories} & \textbf{LLM Category} \\
career, hit, games, season, league, baseball, major\_league\_baseball, signed & \begin{minipage}[t]{\linewidth}
\begin{itemize}
\setlength{\itemsep}{2pt}
\setlength{\parskip}{0pt}
\item Former MLB players
\item Professional baseball facts and figures
\item American baseball league
\end{itemize}
\end{minipage} & Major League Baseball Players and Achievements \\ \addlinespace \midrule
\multicolumn{3}{l}{\textbf{Exemplar Documents}:} \\
\multicolumn{3}{l}{\begin{minipage}[t]{\linewidth}
\begin{itemize}
\setlength{\itemsep}{4pt}
\setlength{\parskip}{0pt}
\item \begin{minipage}[t]{\linewidth}
\textit{\textbf{Doc ID} 2943: $\theta_d=0.61$} \\
\ul{Text:} Brian Wilson ( baseball ) = Brian Patrick Wilson ( born March 16 , 1982 ) is a former American professional baseball relief pitcher .
\end{minipage}
\item \begin{minipage}[t]{\linewidth}
\textit{\textbf{Doc ID} 2928: $\theta_d=0.59$} \\
\ul{Text:} Byron McLaughlin = Byron Scott McLaughlin ( born September 29 , 1955 ) is an American retired professional baseball player , alleged counterfeit [...]
\end{minipage}
\item \begin{minipage}[t]{\linewidth}
\textit{\textbf{Doc ID} 2989: $\theta_d=0.58$} \\
\ul{Text:} Cy Seymour = James Bentley " Cy " Seymour ( December 9 , 1872 – September 20 , 1919 ) was an American center fielder and pitcher in Major League [...]
\end{minipage}
\end{itemize}
\end{minipage}} \\ \addlinespace \midrule
\multicolumn{3}{l}{\textbf{Evaluation Documents}:} \\
\multicolumn{3}{l}{\begin{minipage}[t]{\linewidth}
\begin{itemize}
\setlength{\itemsep}{4pt}
\setlength{\parskip}{0pt}
\item \begin{minipage}[t]{\linewidth}
\textit{\textbf{Doc ID} 2870: $\text{Human Fit}=5.00\pm0.00$, $\text{LLM Fit}=3.97$, $\theta_d=0.74$} \\
\ul{Text:} Jon Lieber = Jonathan Ray Lieber ( born April 2 , 1970 ) is a former Major League Baseball ( MLB ) pitcher .
\end{minipage}
\item \begin{minipage}[t]{\linewidth}
\textit{\textbf{Doc ID} 3001: $\text{Human Fit}=5.00\pm0.00$, $\text{LLM Fit}=4.37$, $\theta_d=0.61$} \\
\ul{Text:} Johnny Evers = John Joseph Evers ( July 21 , 1881 – March 28 , 1947 ) was an American professional baseball second baseman and manager .
\end{minipage}
\item \begin{minipage}[t]{\linewidth}
\textit{\textbf{Doc ID} 2932: $\text{Human Fit}=5.00\pm0.00$, $\text{LLM Fit}=3.22$, $\theta_d=0.49$} \\
\ul{Text:} Bobo Holloman = Alva Lee " Bobo " Holloman ( March 7 , 1923 – May 1 , 1987 ) was an American right-handed pitcher in Major League Baseball who [...]
\end{minipage}
\item \begin{minipage}[t]{\linewidth}
\textit{\textbf{Doc ID} 2865: $\text{Human Fit}=5.00\pm0.00$, $\text{LLM Fit}=5.00$, $\theta_d=0.37$} \\
\ul{Text:} Barry Bonds = Barry Lamar Bonds ( born July 24 , 1964 ) is an American former professional baseball left fielder who played 22 seasons in Major [...]
\end{minipage}
\item \begin{minipage}[t]{\linewidth}
\textit{\textbf{Doc ID} 4300: $\text{Human Fit}=1.33\pm0.47$, $\text{LLM Fit}=1.00$, $\theta_d=0.25$} \\
\ul{Text:} Anthony Davis ( basketball ) = Anthony Marshon Davis , Jr .
\end{minipage}
\item \begin{minipage}[t]{\linewidth}
\textit{\textbf{Doc ID} 5905: $\text{Human Fit}=1.00\pm0.00$, $\text{LLM Fit}=1.00$, $\theta_d=0.12$} \\
\ul{Text:} 1879 Navy Midshipmen football team = The 1879 Navy Midshipmen football team represented the United States Naval Academy in the 1879 college football [...]
\end{minipage}
\item \begin{minipage}[t]{\linewidth}
\textit{\textbf{Doc ID} 10610: $\text{Human Fit}=1.00\pm0.00$, $\text{LLM Fit}=1.00$, $\theta_d=0.00$} \\
\ul{Text:} Edge ( wrestler ) = Adam Joseph Copeland ( born October 30 , 1973 ) is a Canadian actor and retired professional wrestler .
\end{minipage}
\end{itemize}
\end{minipage}} \\
\addlinespace
\bottomrule
\end{tabular}

%% file: tables/topline_table_tau_binarized.tex
\begin{tabular}{@{}p{2.4cm}cccc@{}}
\toprule
\multirow{2}{*}{} & \multicolumn{2}{c}{$\tau\left(\fittauHTM,\cdot \right)$} & \multicolumn{2}{c}{$\tau\left(\ranktauHTM,\cdot \right)$} \\
\cmidrule(lr){2-3}\cmidrule(lr){4-5}
& \texttt{Wiki} & \texttt{Bills} & \texttt{Wiki} & \texttt{Bills} \\
\midrule
NPMI & -0.04 & -0.03 & 0.04 & 0.03 \\
\midrule
\multicolumn{2}{l}{$\textsc{fit/}\ranktauLMTM$} & & & \\
\cmidrule(lr){1-1}
\llamaThreeOneEightB{}   & 0.25 (0.17) & 0.04 (0.16) & -0.01 (0.21) & 0.09 (0.14) \\
\qwenThreeEightB{}     & 0.47 (0.17) & 0.22 (0.19) & 0.44 (0.15) & 0.22 (0.14) \\
\qwenThreeThirtyTwoB{}     & 0.33(0.20) & 0.45 (0.17) & 0.41 (0.15) & 0.48 (0.12) \\
\llamaThreeThreeSeventyB{}  & 0.38 (0.19) & 0.37 (0.15) & 0.36 (0.20) & 0.32 (0.16) \\
\qwenTwoFiveSeventyTwoB{}   & 0.59 (0.13) & 0.56 (0.12) & 0.40 (0.19) & 0.25 (0.16) \\
\gptFourO{}         & 0.60 (0.15) & 0.59 (0.15) & 0.26 (0.21) & 0.26 (0.16) \\
\midrule
\multicolumn{2}{l}{$\textsc{fit/}\ranktauHTM$} & & & \\
\cmidrule(lr){1-1}
\emph{Human} & 0.40 (0.08) & 0.33 (0.08) & 0.14 (0.12) & 0.20 (0.11) \\
\bottomrule
\end{tabular}